\documentclass[11pt]{article}

\usepackage[final]{acl}

\usepackage{times}
\usepackage{latexsym}
\usepackage[T1]{fontenc}
\usepackage[utf8]{inputenc}
\usepackage{microtype}
\usepackage{inconsolata}
\usepackage{fontawesome5}

\definecolor{VistaLink}{HTML}{0171BC}   
\definecolor{VistaIcon}{HTML}{1F2937}   
\definecolor{VistaCite}{HTML}{6B7280}   

\definecolor{darkblue}{rgb}{0, 0, 0.5}
\hypersetup{
    colorlinks=true,
    citecolor=darkblue,
    linkcolor=darkblue,
    urlcolor=darkblue
}
\DeclareRobustCommand{\linkbadge}[3]{%
    \href{#2}{%
        {\textcolor{VistaIcon}{#1}}~%
        {\textcolor{darkblue}{{#3}}}%
    }%
}
\newcommand{\hficon}{%
	\raisebox{-0.15em}{\includegraphics[height=1.25em]{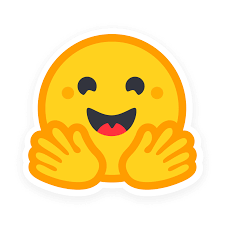}}%
}
\usepackage{amssymb}                              
\usepackage{arydshln}
\usepackage{booktabs}                             
\usepackage{caption}                              
\usepackage{colortbl}                             
\usepackage{graphicx}                             
\usepackage{makecell}                             
\usepackage{mathtools}                            
\usepackage{multirow}                             
\usepackage{pifont}                               
\usepackage{subcaption}                           
\usepackage{tabularx}                             
\usepackage{xcolor}                               
\usepackage{xspace}                               

\newcommand{\cmark}{\ding{51}}
\newcommand{\xmark}{\ding{55}}

\definecolor{darkred}{rgb}{0.6,0.16,0.25}
\definecolor{tabfirst}{rgb}{1,0.7,0.7}
\definecolor{tabsecond}{rgb}{1,0.85,0.7}
\definecolor{tabthird}{rgb}{1,1,0.7}

\newcommand{\modelname}{VISTA\xspace}

\usepackage{placeins}

\begin{document}

\title{VISTA: View-Consistent Self-Verified Training for GUI Grounding}
\author{
	Xinyu Qiu$^{1,2}$\hspace{0.45em}
	Yunzhu Zhang$^{2}$\hspace{0.45em}
	Heng Jia$^{1}$\hspace{0.45em}
	Shuheng Shen$^{2}$\thanks{Corresponding Authors.}
	Changhua Meng$^{2}$\hspace{0.45em}
	Linchao Zhu$^{1}$\textsuperscript{*}
	\\
	\normalfont
	$^{1}$Zhejiang University \quad $^{2}$Venus Team, Ant Group
	\\
    \vspace{0.6em}
    \linkbadge{\faIcon{home}}{https://zjuscl.github.io/VISTA}{Project Page}
    \quad
    \linkbadge{\faIcon{github}}{https://github.com/ZJUSCL/VISTA}{Code}
    \quad
    \linkbadge{\hficon}{https://huggingface.co/inclusionAI/VISTA-4B}{VISTA-4B}
    \quad
    \linkbadge{\hficon}{https://huggingface.co/inclusionAI/VISTA-9B}{VISTA-9B}
}

\maketitle
\begin{abstract}
	When applying Group Relative Policy Optimization (GRPO) for GUI Grounding, rollouts are sampled from a single screenshot view; groups often become either all failures on difficult instances or all successes on easy ones, yielding no useful relative advantage.
	We propose \textbf{\modelname} (\textbf{View-Consistent Self-Verified Training}), a GRPO-based training framework that constructs each comparison group from multiple target-preserving views of the same GUI instance.
	Each view is generated by a crop that keeps the target element visible and remaps its box exactly, so model rollouts are compared across semantically equivalent but geometrically different inputs.
	To stabilize short coordinate generation without turning reinforcement learning into unconditional imitation, \modelname further adds a self-verified cross-view anchor: an oracle answer optimized with an advantage-weighted loss, excluded from the group baseline and activated only when the model has produced a maximum-reward rollout.
	Across five GUI-grounding benchmarks and multiple Qwen backbones, \modelname consistently improves grounding accuracy.
	On ScreenSpot-Pro, it raises Qwen3-VL 4B/8B/30B-A3B from 55.5/52.7/53.7 to 63.4/65.8/67.0.
	Robustness analyses further show higher worst-view accuracy and lower prediction flip rates.
\end{abstract}

\section{Introduction}
GUI grounding enables autonomous agents to interact with digital interfaces by mapping a screenshot and a natural language instruction to a click coordinate~\citep{uitars2, mai-ui, ui_venus, opencua}.
Compared with general visual grounding, GUI grounding is especially sensitive to localization errors, as UI elements such as icons, input fields, and buttons are often small, densely arranged, and visually similar; a slight spatial mistake may activate the wrong element and disrupt the subsequent workflow~\citep{mllmbasedguiagents, guiagents}.

Recent work has made significant progress in applying Group Relative Policy Optimization (GRPO) with verifiable rewards for GUI grounding, where click correctness can be evaluated by a rule-based \emph{point-in-box} reward~\citep{GTA1, gui_r1, GUI_G2}.

\begin{figure*}[t]
	\centering
	\includegraphics[width=\textwidth]{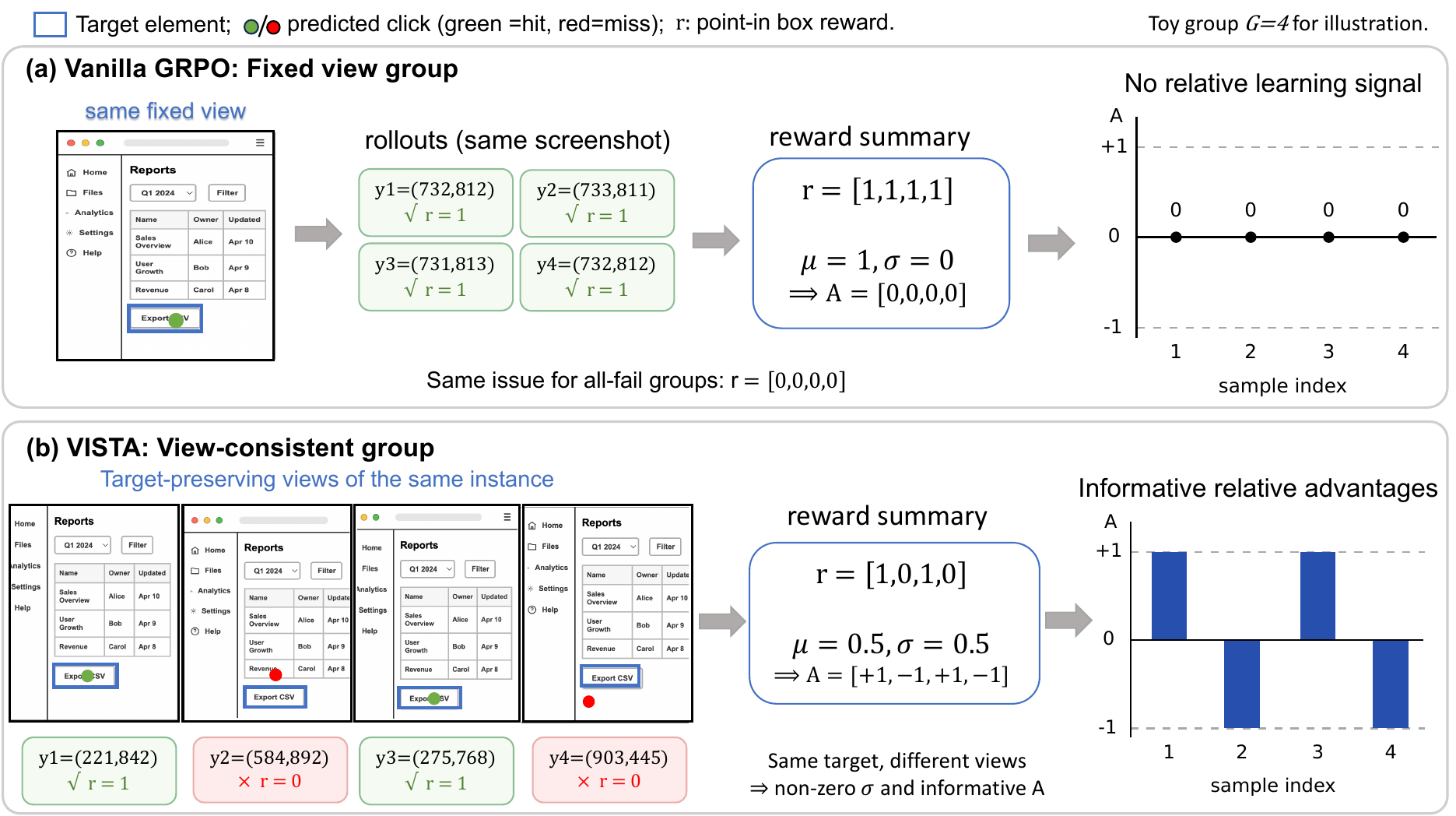}
	\caption{\textbf{Motivation of \modelname.}
In vanilla GRPO, multiple rollouts from the same screenshot can produce homogeneous rewards, yielding zero relative advantage.
\modelname constructs the group from target-preserving views of the same GUI instance.
These views preserve the instruction and target semantics while changing the screenshot geometry.
As a result, \modelname turns homogeneous fixed-view rewards into informative cross-view variation.
}
	\label{fig:reason_prediction_instability}
	\vspace{-1.5em}
\end{figure*}

However, as illustrated in Figure~\ref{fig:reason_prediction_instability},
directly applying GRPO to GUI grounding exposes two forms of reward degeneracy, both of which collapse the group-relative advantage and remove the learning signal~\citep{grpo, dapo,ui_r1}.
With sparse binary \emph{point in box} rewards, repeated rollouts from a single view may all miss the target on difficult screenshots, forming a \emph{fixed view all fail group}.
On easier screenshots, all rollouts may hit the target, which also eliminates reward variance.
Figure~\ref{fig:informative_ratio} shows that fewer than $5\%$ of fixed-view training samples form informative groups, namely groups that are neither all-zero nor all-one.
This issue is severe in GUI grounding because coordinates are tied to the screenshot geometry: a prediction that is correct in one view may shift after a target preserving crop, even when the target remains visible (see Table~\ref{tab:view_setting_comparison})~\citep{mvp}.
Thus, both all-fail and all-correct degeneracy reveal a shared bottleneck: vanilla GRPO constructs groups by repeatedly sampling from one fixed view, which often fails to expose informative differences among rollouts.

These observations suggest that the key design choice is not only how to reward a click, but also how to construct the comparison group.
For GUI grounding, an informative group should preserve the same instruction and target semantics while varying the view geometry, so that the policy is compared across semantically equivalent but geometrically different inputs.
Such view consistent grouping increases the chance of observing a successful rollout on difficult instances and reveals unstable predictions on easy ones.
By keeping the instruction and target unchanged while varying the view geometry, the policy is compared across inputs that require the same action but different coordinate predictions.
Since coordinate generation is sensitive to format and geometry, we separate the model generated rollouts used for group statistics from the oracle answer, which is used only as a conditional stabilizing signal after the model has already succeeded.

In this work, we introduce {\modelname} (\textbf{View-Consistent Self-Verified Training}), a GRPO-based training framework for GUI grounding that revisits group construction from this perspective.
\modelname consists of two components: \textbf{View-Consistent Group Rollout} and \textbf{Self-Verified Cross-View Anchoring}.
View-Consistent Group Rollout constructs each GRPO group from multiple target-preserving views of the same GUI instance, rather than from repeated completions on a single screenshot.
As shown in Figure~\ref{fig:informative_ratio}, this construction increases the fraction of informative groups, namely groups that are neither all-zero nor all-one, from fewer than $5\%$ under fixed-view GRPO to around $20\%$, restoring useful intra-group reward variance for both difficult and easy instances.
Thus, \modelname mitigates all-fail degeneracy primarily through view-consistent group construction, not through unconditional oracle injection.

However, multi-view rollout alone can make coordinate generation unstable on hard or ambiguous views.
To stabilize training, we further introduce a self-verified cross-view anchor that uses oracle answers only as a conditional stabilizing signal, while keeping the GRPO comparison statistics defined by model-generated rollouts.

Our contributions are summarized as follows:
\begin{itemize}
	\item We propose view-consistent group rollout, which constructs GRPO groups from multiple target-preserving views of the same GUI grounding instance and computes model-only group statistics across these views.
	\item We introduce a self-verified cross-view anchor that provides an oracle coordinate only when the current policy has already produced a maximum-reward rollout, preventing oracle targets from changing the GRPO baseline or supervising all-fail groups.
	\item We provide comprehensive validation across model family and benchmarks. On ScreenSpot-Pro, \modelname improves Qwen3-VL series models from 55.5/52.7/53.7 to 63.4/65.8/67.0 at the 4B/8B/30B-A3B scales.
	      On Qwen3.5 initialized backbones, \modelname improves ScreenSpot-Pro over standard GRPO by +2.0/+0.9/+1.2 points at the 4B/9B/35B-A3B scales, with the 35B-A3B model reaching 72.9.
\end{itemize}

\section{Related Work}
\paragraph{GUI Grounding} Recent advances in GUI agents have been driven by specialized GUI grounding models that map natural-language instructions to precise screen coordinates~\cite{mllmbasedguiagents,brainedgui,gpt4twebagent,guiagents,guiagentssurvey}. Early works established the task through large-scale supervised fine-tuning (SFT) on GUI datasets, demonstrating effectiveness across mobile, web, and desktop interfaces~\cite{seeclick,showui, uitars, ariaui, uground, osaltas_and_screenspot_v2, gui_odyssey, aguvis}. More recently, reinforcement learning with rule-based click rewards has emerged as a promising technique to further enhance grounding accuracy beyond SFT baselines~\cite{GTA1, GUI_G2, GUIG1, gui_r1,ui_venus,ui_venus_1_5,qiu2026unified}. However, existing RL approaches suffer from diminishing training effectiveness as optimization progresses.

\paragraph{Oracle Guidance in GRPO Training} 

GRPO has been widely adopted for post-training large language models, consistently improving performance across diverse tasks~\cite{dapo, grpo, grpo_also}. However, GRPO degenerates when group rollouts are uniformly correct or incorrect, collapsing the relative advantage to zero and eliminating the learning signal. To address this, recent works in math and reasoning try to introduce oracle guidance to ensure informative training updates. LUFFY~\cite{lufy} mixes off-policy traces from a stronger teacher into the group, using regularized importance sampling to balance imitation and on-policy exploration. BREAD~\cite{bread} adaptively inserts partial expert prefixes whenever on-policy rollouts fail, guaranteeing at least one successful trace per update. 

\section{Method}
\label{sec:method}
\begin{figure*}[t]
	\vspace{-1.5em}
	\centering
	\includegraphics[width=\textwidth]{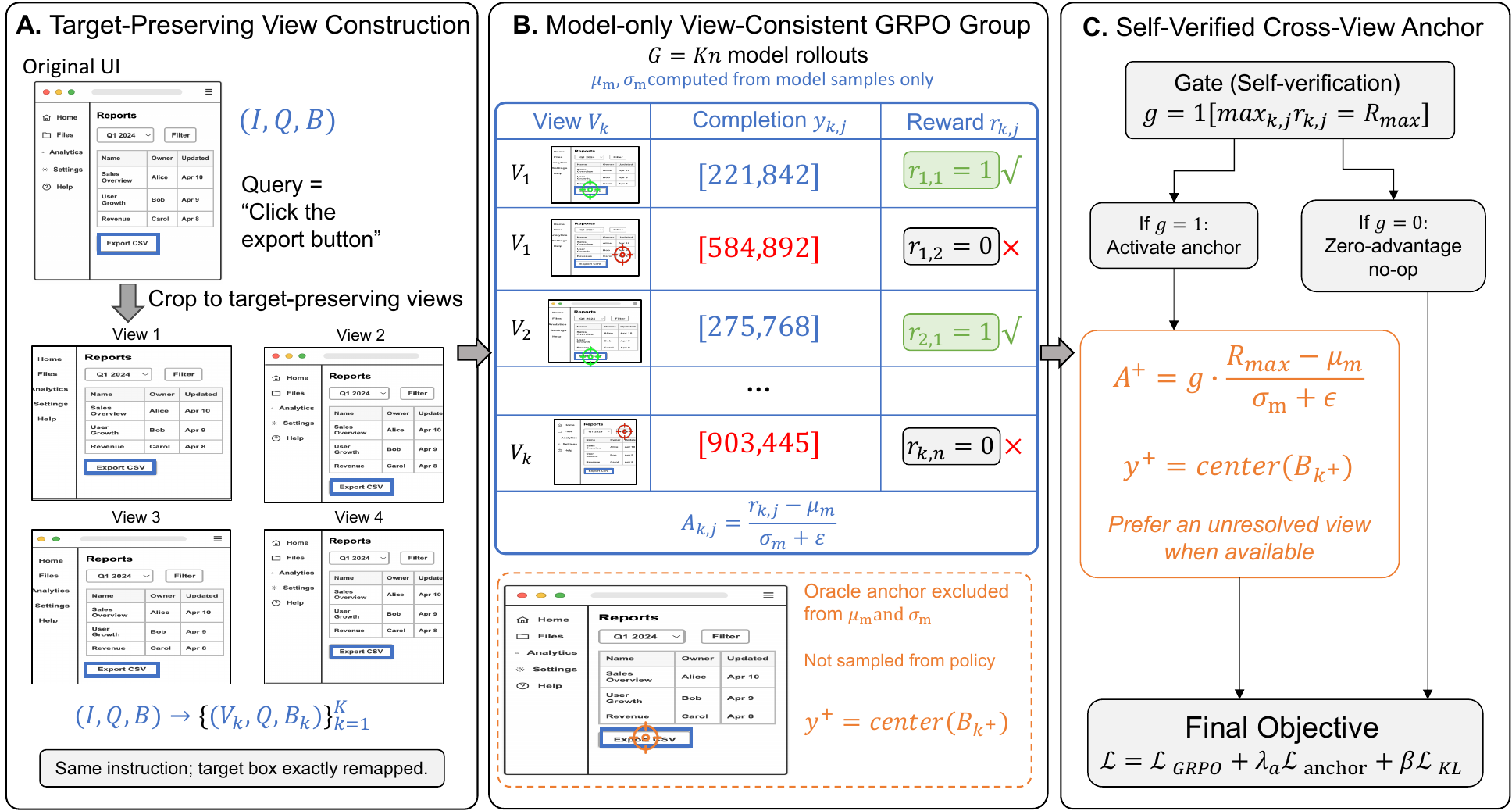}
	\caption{\textbf{Overview of \modelname.}
		\modelname constructs GRPO groups from target-preserving views of the same GUI grounding instance.
		Model rollouts define the group statistics, while an oracle-format center-point anchor is activated only for self-verified groups.
	}
	\label{fig:method_overview}
	\vspace{-1.5em}
\end{figure*}

As illustrated in Figure~\ref{fig:method_overview},
we introduce \modelname, a \textbf{View-Consistent Self-Verified Training} framework for GUI grounding built on GRPO.
The key idea is to construct the GRPO group from multiple target-preserving views of the same screenshot, rather than repeated rollouts on a single fixed rendering.
To stabilize training, \modelname appends an oracle answer as a self-verified cross-view anchor.

\subsection{Problem Setup}

Given a screenshot $I$, an instruction $q$, and a ground-truth target box
$B=(x_1,y_1,x_2,y_2)$, GUI grounding asks a policy $\pi_\theta$ to output a coordinate string $y$.
Following the Qwen-style grounding interface, the output string $y$ is parsed into a click point
$\hat{\mathbf{p}}=(\hat{x},\hat{y})$ in the discrete $0$--$1000$ image coordinate space.
A prediction is considered correct if the parsed click point lies inside the target box.

We use a verifiable reward that matches this click-based interface.
Let $\mathrm{ValidFmt}(y)$ indicate whether $y$ can be parsed as a coordinate string of the form \texttt{[}$x$\texttt{,}$y$\texttt{]}.
The reward is defined as
\begin{equation}
	\small
	R(y,B)=
	\mathbb{I}\!\left[\mathrm{ValidFmt}(y)\right]
	\cdot
	\mathbb{I}\!\left[\hat{\mathbf{p}}\in B\right].
\end{equation}
The maximum reward is therefore $R_{\max}=1$.

\subsection{GRPO for GUI Grounding}

Standard GRPO~\citep{grpo} samples $G$ completions
$\{y_i\}_{i=1}^{G}$ for the same prompt and computes normalized advantages from the group rewards:
\begin{equation}
	\small
	\begin{aligned}
		\mu_G
		 & = \frac{1}{G}\sum_{i=1}^{G} r_i,
		\sigma_G
		= \sqrt{\frac{1}{G}\sum_{i=1}^{G}(r_i-\mu_G)^2}, \\
		A_i
		 & = \frac{r_i-\mu_G}{\sigma_G+\epsilon}.
	\end{aligned}
\end{equation}

Directly applying GRPO to GUI grounding has two limitations: rollouts from a fixed screenshot do not encourage robustness to view changes, and short coordinate outputs often collapse to near-identical strings, producing low reward variance.
We therefore construct each group from target-preserving views of the same task, so that reward comparisons measure localization across renderings rather than repeated completions under one fixed screenshot.

\subsection{View-Consistent Group Rollout}
A valid view transformation must preserve the target geometry and allow exact label remapping.
We use target-preserving crops that fully contain the target, with constrained crop scale and contextual margins to reduce semantic ambiguity.
Each crop is therefore a view-consistent approximation of the original grounding task.

We view each crop as a sample from a conditional view distribution $\mathcal{T}(\cdot \mid I,B)$, so the group baseline is computed over target-preserving renderings of the same grounding instance rather than a single fixed prompt.
With probability $p_{\mathrm{crop}}$, a training instance is converted into $K$ target-preserving cropped views.
Otherwise, we use the original full-screen image as a single view and sample $G$ completions from it, recovering the fixed-view GRPO group.

\paragraph{Target-preserving crop generation.}
Let the original image size be $W\times H$, and let
$B^{\mathrm{px}}=(x_1,y_1,x_2,y_2)$ denote the target box in pixel coordinates.
For each cropped view $k$, the initial crop size is set to
$(w_k,h_k)=(0.9W,\,0.9H)$.
If the resulting crop is smaller than the target box, we enlarge it so that the target can be fully contained.

A valid crop window $C_k=(l_k,t_k,w_k,h_k)$ must satisfy
$B^{\mathrm{px}}\subseteq C_k$.
For each view, we independently sample the top-left corner uniformly from these feasible ranges:
\begin{equation}
	\small
	\begin{aligned}
		l_k & \sim \mathcal{U}
		\left(
		\max(0,x_2-w_k),\,
		\min(x_1,W-w_k)
		\right),               \\
		t_k & \sim \mathcal{U}
		\left(
		\max(0,y_2-h_k),\,
		\min(y_1,H-h_k)
		\right),
	\end{aligned}
\end{equation}
for $k=1,\ldots,K$.
This yields $K$ cropped views without truncating the target UI element.

\paragraph{Coordinate remapping.}
For each valid crop, we remap the target box into the cropped coordinate frame and normalize it:
\begin{equation}
	\small
	\begin{aligned}
		B_k=\bigl(
		 & 1000\tfrac{x_1-l_k}{w_k},
		1000\tfrac{y_1-t_k}{h_k},    \\
		 & 1000\tfrac{x_2-l_k}{w_k},
		1000\tfrac{y_2-t_k}{h_k}
		\bigr).
	\end{aligned}
\end{equation}

The cropped image is then resized for the VLM input, while the target box is obtained by exact geometric remapping.

\paragraph{Group construction.}
Given $K$ target-preserving views
$\{(V_k,B_k)\}_{k=1}^{K}$, we sample $n=G/K$ completions per view, assuming $K$ divides the GRPO group size $G$:
\begin{equation}
	\small
	\begin{aligned}
		y_{k,j}
		 & \sim
		\pi_{\theta_{\mathrm{old}}}(\cdot \mid V_k,q), \\
		r_{k,j}
		 & = R(y_{k,j},B_k),
	\end{aligned}
	\quad
	k=1,\ldots,K,\;
	j=1,\ldots,n .
\end{equation}
The model-generated group contains $G=Kn$ rollouts.
Its statistics are computed as
\begin{equation}
	\small
	\begin{aligned}
		\mu_{\mathrm{m}}
		 & =
		\frac{1}{G}
		\sum_{k=1}^{K}\sum_{j=1}^{n} r_{k,j}, \\
		\sigma_{\mathrm{m}}
		 & =
		\sqrt{
			\frac{1}{G}
			\sum_{k=1}^{K}\sum_{j=1}^{n}
			(r_{k,j}-\mu_{\mathrm{m}})^2
		},
	\end{aligned}
\end{equation}
and the corresponding model-sample advantages are
\begin{equation}
	A_{k,j}
	=
	\frac{r_{k,j}-\mu_{\mathrm{m}}}
	{\sigma_{\mathrm{m}}+\epsilon}.
\end{equation}

When $n>1$, the group retains within-view sampling stochasticity; when $K=G$, it maximizes cross-view diversity.
In both cases, advantages compare predictions across views of the same semantic target.

\subsection{Self-Verified Cross-View Anchoring}
As shown by the training diagnostics in Figure~\ref{fig:training_diagnostics},
multi-view rollouts improve exploration over target-preserving views, but they can also make the reward signal less stable.
In particular, GUI grounding outputs are short coordinate strings, and repeated negative advantages on invalid or poorly localized completions may reduce the probability of producing valid coordinates.
To stabilize training without reverting to unconditional supervised fine-tuning, we introduce a \textbf{self-verified cross-view anchor}.
We use the term {self-verified} to indicate that the anchor is activated only when a model-generated rollout, rather than the oracle sequence itself, achieves the maximum verifier reward.

The anchor uses the ground-truth geometry only when the current policy has already produced evidence that the instance is solvable within the same view-consistent group.
A maximum-reward rollout indicates that the policy can ground the target in at least one target-preserving view.
In contrast, if no rollout reaches the maximum reward, the group provides no model-side evidence that the instance is currently solvable.
In such cases, forcing an oracle coordinate would turn the update into unconditional supervised learning.
We therefore activate the anchor only for self-verified groups.

\subsection{Training Objective}

\begin{table*}[!t]
	\vspace{-1em}
	\centering
	\scriptsize
	\setlength{\tabcolsep}{3pt}
	\renewcommand{\arraystretch}{0.95}
	\caption{
		\textbf{Overall results of \modelname\ across five GUI-grounding benchmarks.}
		We highlight the \colorbox{tabfirst}{best} and \colorbox{tabsecond}{second-best} results within each model size category.
		$^*$ denotes our evaluated results.
	}
	\label{tab:overview_results}
	\begin{tabularx}{\textwidth}{
			>{\raggedright\arraybackslash\hsize=2.2\hsize\linewidth=\hsize}X
			*{6}{>{\centering\arraybackslash\hsize=0.8\hsize\linewidth=\hsize}X}}
		\toprule
		\textbf{Model}                    & \textbf{SSPro}             & \textbf{SSV2}              & \textbf{MMBench-L2}        & \textbf{OSWorld-G-R}       & \textbf{OSWorld-G}         & \textbf{Avg.}              \\
		\midrule
		\multicolumn{7}{l}{\color{darkred}{\textit{$\approx$4B}}}                                                                                                                                                       \\
		Qwen3-VL-4B$^*$ \citep{Qwen3-VL}   & 55.5                       & 88.5                       & 85.3                       & 67.9                       & 58.2                       & 71.1                       \\
		Step-GUI-4B \citep{step-gui}      & 60.0                       & 93.6                       & 84.0                       & 66.9                       & 60.5                       & 73.0                       \\
		\rowcolor{gray!10}
		{\modelname-4B}                   & \colorbox{tabsecond}{63.4} & \colorbox{tabsecond}{94.4} & \colorbox{tabsecond}{86.7} & \colorbox{tabsecond}{69.4} & \colorbox{tabsecond}{63.8} & \colorbox{tabsecond}{75.5} \\
		\rowcolor{gray!10}
		\textit{\quad + MVP}                    & \colorbox{tabfirst}{71.6}  & \colorbox{tabfirst}{94.5}  & \colorbox{tabfirst}{86.8}  & \colorbox{tabfirst}{69.6}  & \colorbox{tabfirst}{64.0}  & \colorbox{tabfirst}{77.3}  \\
		\midrule
		\multicolumn{7}{l}{\color{darkred}{\textit{$\approx$8B}}}                                                                                                                                                       \\
		UI-TARS-1.5-7B \citep{uitars}     & 35.7                       & 91.6                       & 64.3                       & 64.2                       & 52.8                       & 61.7                       \\
		OpenCUA-7B \citep{opencua}        & 50.0                       & 92.3                       & -                          & -                          & 55.3                       & -                          \\
		GTA1-7B \citep{GTA1}              & 50.1                       & 92.4                       & -                          & 67.7                       & 60.1                       & -                          \\
		UI-Venus-7B \citep{ui_venus}      & 50.8                       & 94.1                       & 79.9                       & 61.7                       & 54.6                       & 68.2                       \\
		GUI-Owl-7B \citep{mobileagentv3}  & 54.9                       & 92.8                       & 80.5                       & -                          & 55.9                       & -                          \\
		Step-GUI-8B \citep{step-gui}      & 62.6                       & 95.1                       & 85.6                       & 70.0                       & -                          & -                          \\
		Qwen3-VL-8B$^*$ \citep{Qwen3-VL}   & 52.7                       & 91.7                       & 81.3                       & 64.4                       & 54.8                       & 69.0                       \\
		Holo2-8B \citep{holo2}            & 58.9                       & 93.2                       & 84.5                       & 70.1                       & \colorbox{tabfirst}{63.5}  & 74.0                       \\
		MAI-UI-8B \citep{mai-ui}          & \colorbox{tabsecond}{65.8} & 95.2                       & \colorbox{tabfirst}{88.8}  & 68.6                       & 60.1                       & 75.7                       \\
		\rowcolor{gray!10}
		{\modelname-8B}                   & \colorbox{tabsecond}{65.8} & \colorbox{tabsecond}{95.5} & 86.8                       & \colorbox{tabsecond}{70.8} & 62.4                       & \colorbox{tabsecond}{76.3} \\
		\rowcolor{gray!10}
		\textit{\quad + MVP}                    & \colorbox{tabfirst}{72.0}  & \colorbox{tabfirst}{95.6}  & \colorbox{tabsecond}{87.3} & \colorbox{tabfirst}{70.9}  & \colorbox{tabsecond}{63.1} & \colorbox{tabfirst}{77.8}  \\
		\midrule
		\multicolumn{7}{l}{\color{darkred}{\textit{$\geq$30B}}}                                                                                                                                                         \\
		Qwen3-VL-30B-A3B$^*$ \citep{Qwen3-VL} & 53.7                    & 94.7                       & 83.7                       & 69.3                       & 66.5                       & 73.6                       \\
		Holo2-30B-A3B \citep{holo2}       & 66.1                       & 94.9                       & \colorbox{tabsecond}{86.8} & \colorbox{tabfirst}{76.1}  & 65.2                       & \colorbox{tabsecond}{77.8} \\
		OpenCUA-32B \citep{opencua}       & 55.3                       & 93.4                       & -                          & 70.2                       & 59.6                       & -                          \\
		GUI-Owl-32B \citep{mobileagentv3} & 58.0                       & 93.1                       & 83.0                       & -                          & 58.0                       & -                          \\
		GTA1-32B \citep{GTA1}             & 63.6                       & 95.2                       & -                          & 72.2                       & 65.2                       & -                          \\
		OpenCUA-72B \citep{opencua}       & 60.8                       & 92.9                       & -                          & -                          & -                          & -                          \\
		UI-Venus-72B \citep{ui_venus}     & 61.9                       & \colorbox{tabsecond}{95.3} & 86.3                       & 69.5                       & 62.2                       & 75.0                       \\
		\rowcolor{gray!10}
		{\modelname-30A3B}                & \colorbox{tabsecond}{67.0} & 95.2                       & \colorbox{tabsecond}{86.8} & 72.0                       & \colorbox{tabsecond}{67.1} & 77.6                       \\
		\rowcolor{gray!10}
		\textit{\quad + MVP}                    & \colorbox{tabfirst}{74.1}  & \colorbox{tabfirst}{95.4}  & \colorbox{tabfirst}{87.6}  & \colorbox{tabsecond}{72.5} & \colorbox{tabfirst}{67.6}  & \colorbox{tabfirst}{79.4}  \\
		\bottomrule
	\end{tabularx}
	\vspace{-1.5em}
\end{table*}

Let $\mathcal{P}=\{(k,j): r_{k,j}=R_{\max}\}$ denote the set of maximum-reward model rollouts, and let $\mathcal{N}$ be its complement within the group.
When $\mathcal{N}$ is non-empty, we choose the anchor view uniformly from $\mathcal{N}$, so that the oracle anchor preferentially targets a view where the current policy has not yet produced a perfect prediction.
If all model rollouts are already perfect, we choose the anchor view uniformly from the full group.

Given the selected view $V_{k^{+}}$ and its remapped box $B_{k^{+}}=(x_1,y_1,x_2,y_2)$, we construct the oracle-format coordinate sequence as the box center,
$y^{+}=\texttt{[}\lfloor(x_1+x_2)/2\rfloor\texttt{, }\lfloor(y_1+y_2)/2\rfloor\texttt{]}$.

Although this sequence is geometrically valid by construction, it is not assigned a positive reward unconditionally.
The anchor advantage is computed with the model-only group baseline:
\begin{equation}
	A^{+}
	=
	\mathbb{I}[\mathcal{P}\neq\emptyset]\,
	\frac{R_{\max}-\mu_{\mathrm{m}}}{\sigma_{\mathrm{m}}+\epsilon}.
\end{equation}

The oracle sequence is excluded from $\mu_{\mathrm{m}}$ and $\sigma_{\mathrm{m}}$.
This is important: including the oracle in the group statistics would allow a ground-truth sequence to shift the baseline, creating implicit supervision even for groups where the model never succeeds.
By contrast, our model-only baseline ensures that the anchor contributes a nonzero supervised signal only when the group is self-verified by at least one maximum-reward rollout.
When all model rollouts fail to reach $R_{\max}$, the anchor becomes a zero-advantage no-op.
The model-generated rollouts may still contribute standard GRPO updates if their rewards are non-identical, but no oracle-anchor update is applied.

This gating prevents the anchor from becoming unconditional supervised fine-tuning on groups where the policy has not yet produced any evidence of successful grounding.
It also reduces the influence of noisy annotations or ambiguous target views.

This design distinguishes \modelname from common mixed-policy or off-policy-supervised variants~\citep{lufy, bread}.
Ground-truth injection inserts oracle completions into the RL update or candidate group, which can alter the effective comparison set.
Fixed-weight SFT mixing adds a supervised loss with a constant coefficient, regardless of whether the current policy has already discovered the target.
In contrast, our anchor is self-verified, baseline-excluded, and conditional: it provides a supervised stabilizer only for groups where the current policy has already demonstrated successful grounding, while avoiding unconditional imitation on groups where grounding has not yet emerged.

For a model-generated completion $y_{k,j}$, we use the clipped GRPO objective
\begin{equation}
	\small
	\begin{aligned}
		\ell_{\mathrm{clip}}(y_{k,j},A_{k,j})
		 & =
		\sum_t
		\min\left(
		\rho_{k,j,t}A_{k,j},
		\bar{\rho}_{k,j,t}A_{k,j}
		\right), \\
		\bar{\rho}_{k,j,t}
		 & =
		\mathrm{clip}
		\left(
		\rho_{k,j,t},
		1-\epsilon_{\mathrm{clip}},
		1+\epsilon_{\mathrm{clip}}
		\right),
	\end{aligned}
\end{equation}
where
\begin{equation}
	\small
	\rho_{k,j,t}
	=
	\frac{
		\pi_{\theta}
		(y_{k,j,t}\mid V_k,q,y_{k,j,<t})
	}{
		\pi_{\theta_{\mathrm{old}}}
		(y_{k,j,t}\mid V_k,q,y_{k,j,<t})
	}.
\end{equation}

The oracle sequence is not an on-policy sample.
We therefore optimize it as an advantage-weighted supervised anchor:
\begin{equation}
	\small
	\ell_{\mathrm{anchor}}(y^{+},A^{+})
	=
	\mathrm{sg}(A^{+})
	\sum_t
	\log
	\pi_{\theta}
	\left(
	y^{+}_t
	\mid
	V_{k^{+}},q,y^{+}_{<t}
	\right),
\end{equation}
where $\mathrm{sg}(\cdot)$ denotes stop-gradient.

The final loss is
\begin{equation}
	\small
	\begin{aligned}
		\mathcal{L}_{\text{\modelname}}
		=
		- & \frac{1}{G+1}
		\Bigg[
			\sum_{k=1}^{K}
			\sum_{j=1}^{n}
		\ell_{\mathrm{clip}}(y_{k,j},A_{k,j}) \\
		  & +
			\lambda_{\mathrm{a}}
			\ell_{\mathrm{anchor}}(y^{+},A^{+})
			\Bigg]
		+
		\beta\mathcal{L}_{\mathrm{KL}},
	\end{aligned}
\end{equation}
where $\lambda_{\mathrm{a}}$ controls the strength of the anchor loss and is set to $1$ unless otherwise specified.
In implementation, the oracle coordinate sequence is appended as an additional sequence and optimized through the same token-level training pipeline as the model completions.
\section{Experiments}

\begin{figure*}[t]
	\vspace{-1.5em}
	\centering
	\begin{subfigure}[t]{0.32\textwidth}
		\centering
		\includegraphics[width=\linewidth]{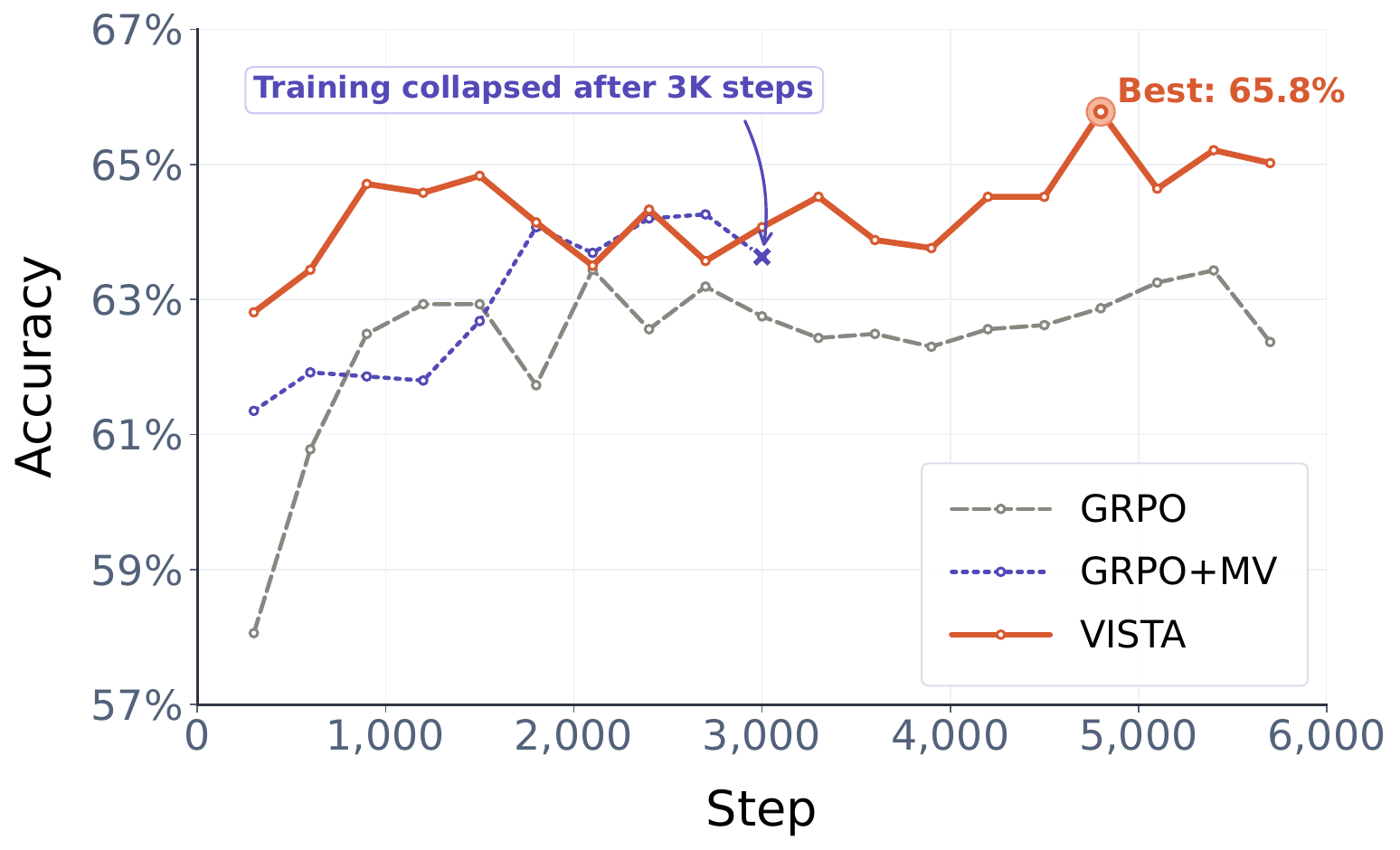}
		\caption{ScreenSpot-Pro accuracy $\uparrow$.}
		\label{fig:sspro_accuracy}
	\end{subfigure}
	\hfill
	\begin{subfigure}[t]{0.32\textwidth}
		\centering
		\includegraphics[width=\linewidth]{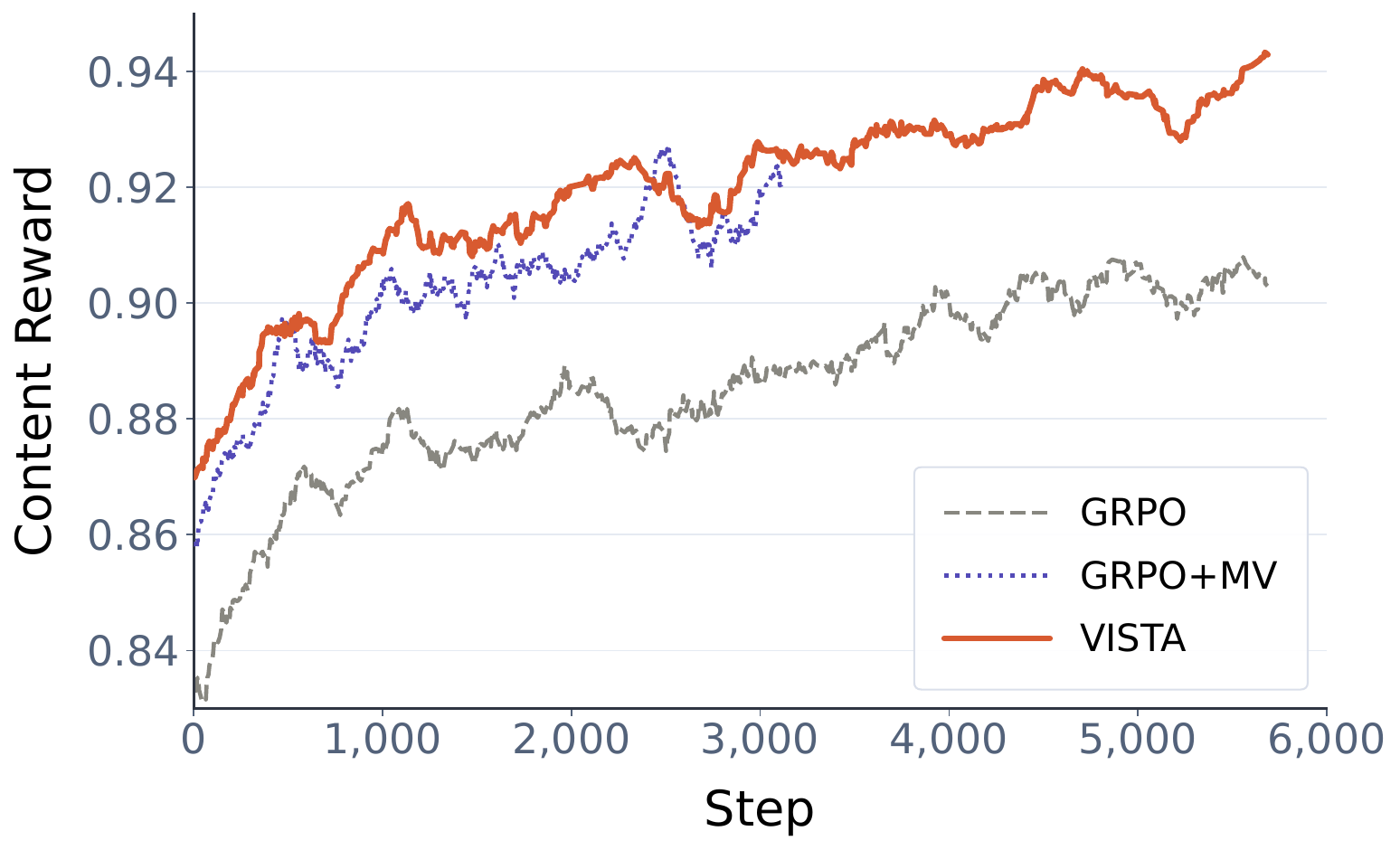}
		\caption{Content reward $\uparrow$.}
		\label{fig:content_reward}
	\end{subfigure}
	\hfill
	\begin{subfigure}[t]{0.32\textwidth}
		\centering
		\includegraphics[width=\linewidth]{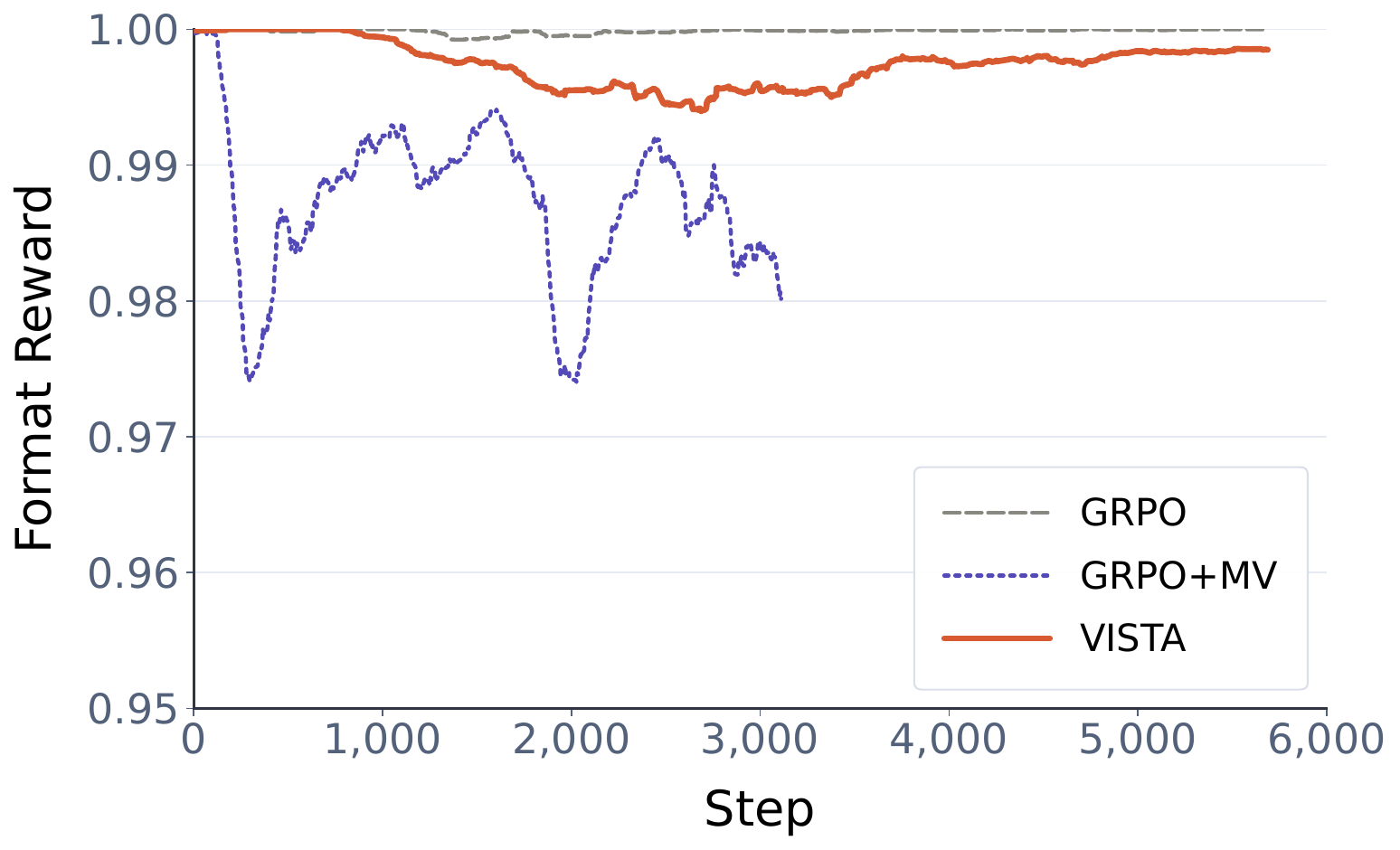}
		\caption{Format reward $\uparrow$.}
		\label{fig:format_reward}
	\end{subfigure}

	\begin{subfigure}[t]{0.32\textwidth}
		\centering
		\includegraphics[width=\linewidth]{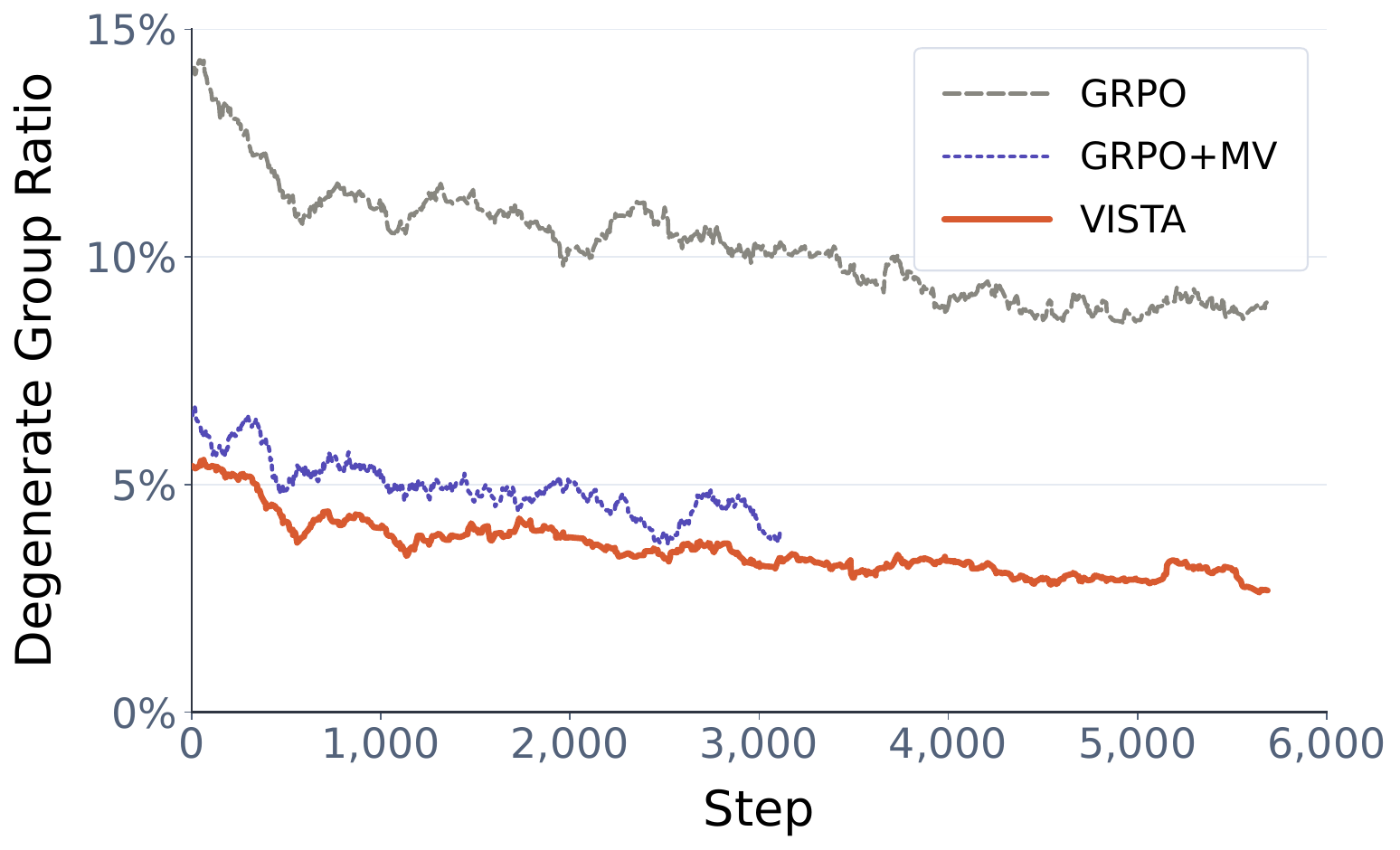}
		\caption{All-zero degenerate ratio $\downarrow$.}
		\label{fig:degenerate_ratio_all0}
	\end{subfigure}
	\hfill
	\begin{subfigure}[t]{0.32\textwidth}
		\centering
		\includegraphics[width=\linewidth]{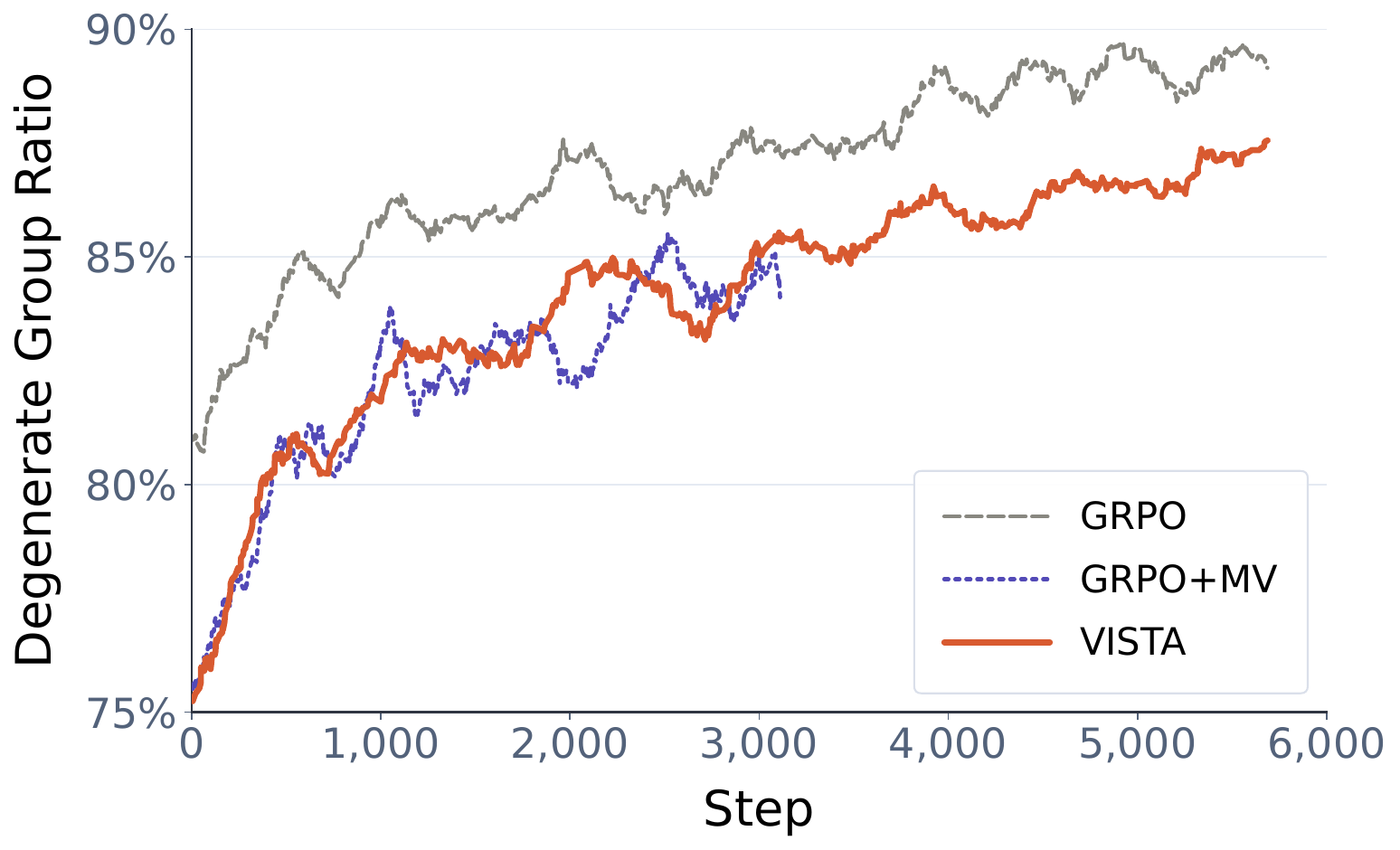}
		\caption{All-one degenerate ratio $\downarrow$.}
		\label{fig:degenerate_ratio_all1}
	\end{subfigure}
	\hfill
	\begin{subfigure}[t]{0.32\textwidth}
		\centering
		\includegraphics[width=\linewidth]{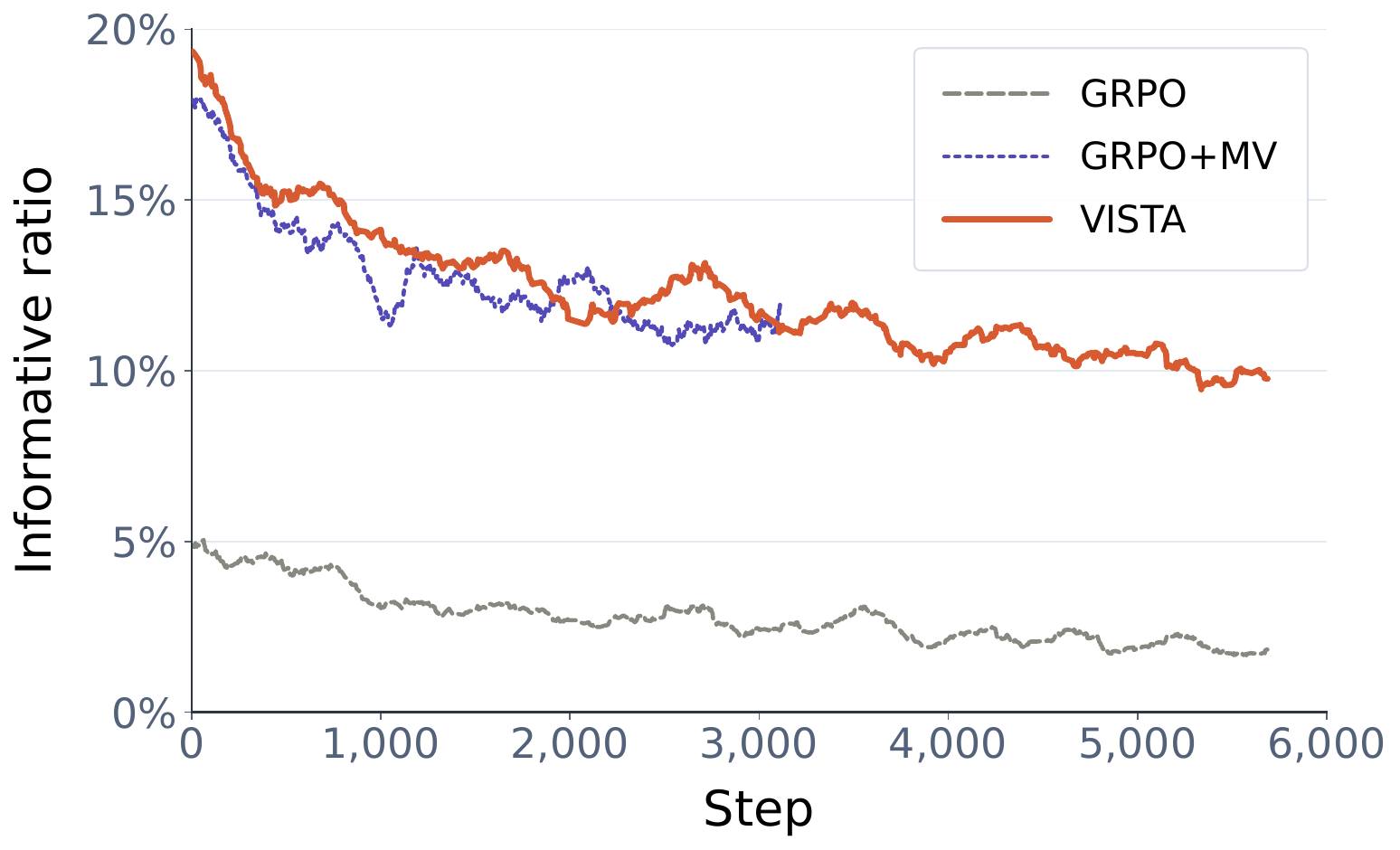}
		\caption{Informative group ratio $\uparrow$.}
		\label{fig:informative_ratio}
	\end{subfigure}
	\caption{\textbf{Training dynamics and reward diagnostics.}
	Degenerate group ratios: “All-zero” groups contain only zero-reward responses, while “All-one” groups contain only full-reward responses; both are uninformative for policy-gradient updates and lower is better. The informative group ratio,
	$1-\mathrm{ratio}_{\mathrm{all-zero}}-\mathrm{ratio}_{\mathrm{all-one}}$,
	is the share of non-degenerate groups containing both successful and unsuccessful rollouts.
	}
	\label{fig:training_diagnostics}
	\vspace{-1.5em}
\end{figure*}

We evaluate \modelname from three perspectives.
First, we test whether view-consistent self-verified training improves GUI grounding across model scales and benchmarks.
Second, we examine whether the improvement is tied to a specific Qwen3-VL initialization.
Third, we isolate the effects of view-consistent dynamic cropping, self-verified anchoring, and crop-set robustness.
Unless otherwise specified, \modelname uses $K{=}G{=}8$ target-preserving views and appends one oracle center-point completion per training block.

\subsection{Experimental Setup}

\paragraph{Benchmarks and metrics.}
We report results on five GUI-grounding benchmarks: ScreenSpot-Pro~\citep{li2025screenspotpro}, ScreenSpot-V2~\citep{osaltas_and_screenspot_v2}, MMBench-GUI L2~\citep{wang2025mmbenchgui}, OSWorld-G-R, and OSWorld-G~\citep{osworld_g}.
These benchmarks cover mobile, web, desktop, and high-resolution professional software interfaces.
For all datasets, the model predicts a normalized coordinate in the $0$--$1000$ frame, and a prediction is counted as correct when the point falls inside the target element.
We use accuracy as the primary metric and report the average over the listed benchmarks when all required numbers are available in the corresponding table.
All evaluations are conducted with deterministic decoding at temperature 0.

\paragraph{Models and baselines.}
Our main experiments instantiate \modelname on Qwen3-VL backbones at the 4B, 8B, and 30B-A3B scales.
We compare against the original Qwen3-VL models and recent GUI grounding models within the same parameter-scale groups.
All \modelname rows use standard single-view inference unless marked with \textit{+ MVP}.
MVP~\citep{mvp} is an orthogonal inference-time multi-view aggregation method; we include it to test whether a model trained with view-consistent rollouts remains compatible with test-time view aggregation.
For cross-backbone evaluation, we additionally train Qwen3.5-initialized models and compare \modelname against standard GRPO under the same backbone family.

\paragraph{Training dataset and cost.}
We train \modelname on roughly 120K GUI-grounding samples curated from open-source datasets, including SeeClick~\citep{seeclick}, Widget Captioning~\citep{li2020widget}, ShowUI-web~\citep{showui}, UI-RefExp~\citep{bai2021uibert}, OmniAct~\citep{omniact}.
Compared with standard GRPO, \modelname incurs a moderate training overhead:
across different model sizes, training for the same number of steps
increases wall-clock time by approximately $25\%$ under the same training setup.

\paragraph{Implementation details.}
All \modelname runs use $G{=}8$ generations.
For \modelname, we use $K{=}8$ target-preserving views, one completion per view, and one self-verified oracle anchor with $\lambda_{\mathrm{a}}{=}1$.
Dynamic cropping is applied to $p_{crop}=80\%$ of training examples, while the remaining $20\%$ use pass-through full-screen views.
We train with DeepSpeed ZeRO-3, bfloat16 precision, frozen vision modules, a learning rate of $1{\times}10^{-6}$, KL coefficient $\beta{=}0.04$, and total batch size 128.

\subsection{Main Results}
Table~\ref{tab:overview_results} shows that \modelname consistently improves the Qwen3-VL backbone across scales.
At 4B, 8B, and 30B-A3B, the average score increases from 71.1/69.0/73.6 to 75.5/76.3/77.6, respectively.
The largest gains appear on ScreenSpot-Pro, where small targets and dense high-resolution interfaces make coordinate errors especially costly: \modelname improves Qwen3-VL by +7.9, +13.1, and +13.3 points at the three scales.
By contrast, ScreenSpot-V2 is already near saturation for strong GUI models, so the gains there are smaller but still positive.
This pattern supports the central motivation of \modelname: view-consistent rollouts and self-verified anchoring mainly help on difficult instances rather than only improving easy benchmarks.

\begin{table}[!t]
	\centering
	\scriptsize
	\setlength{\tabcolsep}{3pt}
	\renewcommand{\arraystretch}{0.95}
	\caption{
		\textbf{Cross-backbone generalization with Qwen3.5-initialized models.}
	}
	\vspace{-1em}
	\label{tab:backbone_generalization}
	\begin{tabularx}{\linewidth}{
			>{\raggedright\arraybackslash\hsize=1.6\hsize\linewidth=\hsize}X
			*{4}{>{\centering\arraybackslash\hsize=0.85\hsize\linewidth=\hsize}X}}
		\toprule
		\textbf{Method}      & \textbf{SSPro} & \textbf{SSV2} & \makecell{\textbf{OS} \textbf{-G}} & \makecell{\textbf{OS}\textbf{-G-R}} \\
		\midrule
		\multicolumn{5}{l}{\color{darkred}{\textit{4B}}}                                                                                 \\
		Qwen3.5-4B           & 60.3           & 90.4          & 54.4                               & 66.8                                \\
		GRPO-4B              & 62.2           & \textbf{94.2} & 59.9                               & 69.2                                \\
		\rowcolor{gray!10}
		{\modelname-4B}      & \textbf{64.2}  & 93.8          & \textbf{61.2}                      & \textbf{69.7}                       \\
		\rowcolor{gray!10}
		$\Delta$     & \textbf{+2.0}  & -0.4          & \textbf{+1.3}                      & \textbf{+0.5}                       \\
		\midrule
		\multicolumn{5}{l}{\color{darkred}{\textit{9B}}}                                                                                 \\
		Qwen3.5-9B           & 65.2           & 91.9          & 63.1                               & 74.6                                \\
		GRPO-9B              & 68.3           & 95.2          & 67.5                               & 75.2                                \\
		\rowcolor{gray!10}
		{\modelname-9B}      & \textbf{69.2}  & \textbf{95.8} & \textbf{68.1}                      & \textbf{75.5}                       \\
		\rowcolor{gray!10}
		$\Delta$     & \textbf{+0.9}  & \textbf{+0.6} & \textbf{+0.6}                      & \textbf{+0.3}                       \\
		\midrule
		\multicolumn{5}{l}{\color{darkred}{\textit{35B-A3B}}}                                                                            \\
		Qwen3.5-35B-A3B      & 68.6           & 93.8          & 65.8                               & 72.5                                \\
		GRPO-35B-A3B         & 71.7           & 95.7          & 70.4                               & 74.3                                \\
		\rowcolor{gray!10}
		{\modelname-35B-A3B} & \textbf{72.9}  & \textbf{95.8} & \textbf{71.5}                      & \textbf{75.3}                       \\
		\rowcolor{gray!10}
		$\Delta$    & \textbf{+1.2}  & \textbf{+0.1} & \textbf{+1.1}                      & \textbf{+1.0}                       \\
		\bottomrule
	\end{tabularx}
	\vspace{-1.5em}
\end{table}


\begin{table}[t]
	\vspace{-1em}
	\centering
	\caption{\textbf{Ablation study on the components of \modelname.}
		The shaded row marks the full model with dynamic cropping and adaptive supervision.
		View group denotes view-consistent rollout groups; Anchor denotes self-verified adaptive anchor supervision.
	}
	\vspace{-0.5em}
	\label{tab:ablation_main}
	\footnotesize
	\setlength{\tabcolsep}{5pt}
	\renewcommand{\arraystretch}{1.08}

	\begin{tabularx}{0.98\linewidth}{@{}>{\raggedright\arraybackslash}Xcccc@{}}
		\toprule
		\multirow{2}{*}{Method}
		              & \multicolumn{2}{c}{Components}
		              & \multicolumn{2}{c}{Accuracy $\uparrow$}                                          \\
		\cmidrule(lr){2-3}
		\cmidrule(l){4-5}
		              & View group                              & Anchor & SSV2          & SSPro         \\
		\midrule
		Qwen3-VL-8B   & \xmark                                  & \xmark & 91.7          & 54.6          \\
		SFT           & \xmark                                  & \xmark & 93.6          & 59.8          \\
		SFT + aug     & \xmark                                  & \xmark & 94.6          & 60.5          \\
		\midrule
		GRPO          & \xmark                                  & \xmark & 95.2          & 63.4          \\
		GRPO + aug    & \xmark                                  & \xmark & 95.1          & 64.0          \\
		GRPO + crop   & \cmark                                  & \xmark & 95.4          & 64.3          \\
		GRPO + anchor & \xmark                                  & \cmark & 95.3          & 64.8          \\
		\rowcolor{gray!10}
		\modelname    & \cmark                                  & \cmark & \textbf{95.5} & \textbf{65.8} \\
		\bottomrule
	\end{tabularx}

	\vspace{0.2em}
	\begin{minipage}{0.98\linewidth}
		\footnotesize
	\end{minipage}
	\vspace{-1em}
\end{table}

\modelname also combines well with inference-time multi-view prediction.
Adding MVP further raises the average scores to 77.3, 77.8, and 79.4 at the 4B, 8B, and 30B-A3B scales.
The improvement is again most pronounced on ScreenSpot-Pro, where the \textit{+ MVP} rows reach 71.6, 72.0, and 74.1.
Since MVP is applied only at inference time, these results indicate that the robustness learned by \modelname is complementary to test-time view aggregation.

Figure~\ref{fig:training_diagnostics} summarizes the training dynamics and reward diagnostics.
Compared with standard GRPO, multi-view rollout reduces all-zero degeneracy and improves content reward,
but its format reward is noticeably unstable during training.
This confirms that view diversification alone can introduce additional coordinate-format instability.
By combining view-consistent rollout with the model-verified anchor, \modelname maintains a high and stable
format reward while improving content reward and ScreenSpot-Pro accuracy.

\begin{table}[t]
    \centering
    \caption{\textbf{Ablation study on anchor supervision} (Qwen3-VL-8B).
    Adaptive gating avoids unverified oracle updates while preserving a model-only baseline.
    Gate denotes self-verified adaptive activation.}
    \vspace{-0.5em}
    \label{tab:ablation_supervision}
    \footnotesize
\newcolumntype{L}{>{\raggedright\arraybackslash}X}
\newcolumntype{C}{>{\centering\arraybackslash}X}
\renewcommand{\arraystretch}{1.08}

\begin{tabularx}{0.92\linewidth}{@{}LCCC@{}}
\toprule
Anchor & Gate & SSV2 & SSPro \\
\midrule
None        & --     & 95.4 & 64.3 \\
Normalized  & \xmark & 93.8 & 57.8 \\
Const. SFT  & \xmark & 94.8 & 63.9 \\
\rowcolor{gray!10}
Normalized  & \cmark & \textbf{95.5} & \textbf{65.8} \\
\bottomrule
\end{tabularx}

\begin{minipage}{0.92\linewidth}
\footnotesize
\end{minipage}
\vspace{-1.5em}

\end{table}

\subsection{Cross-Backbone Generalization}

To test whether \modelname depends on Qwen3-VL initialization, we train Qwen3.5-initialized backbones at 4B, 9B, and 35B-A3B scales and compare them against standard GRPO.
Table~\ref{tab:backbone_generalization} shows that \modelname transfers beyond the Qwen3-VL family.
It improves ScreenSpot-Pro at all three scales and improves OSWorld-G-R for each reported backbone.
At 9B, \modelname still improves ScreenSpot-Pro, ScreenSpot-V2, and OSWorld-G-R.


\begin{table}[t]
    \vspace{-1em}
\centering
\caption{\textbf{Comparison under different view settings.}
Worst denotes worst-view accuracy; VCR is view-consistency rate; Flip is prediction flip rate; Base is Qwen3-VL-8B.}
    \vspace{-0.5em}
\label{tab:view_setting_comparison}
\footnotesize
\renewcommand{\arraystretch}{1.08}

\begin{tabular*}{0.92\linewidth}{@{\extracolsep{\fill}}lccccc@{}}
\toprule
\multirow{2}{*}{Model}
& \multicolumn{3}{c}{Accuracy $\uparrow$}
& \multirow{2}{*}{VCR $\uparrow$}
& \multirow{2}{*}{Flip $\downarrow$} \\
\cmidrule(lr){2-4}
& Orig. & Crop & Worst & & \\
\midrule
Base   & 81.82 & 81.25 & 71.46 & 75.76 & 17.28 \\
GRPO   & 94.19 & 93.00 & 87.63 & 88.38 & 8.31  \\
\rowcolor{gray!10}
\modelname & \textbf{95.71} & \textbf{96.25} & \textbf{92.42}
       & \textbf{90.40} & \textbf{5.80} \\
\bottomrule
\end{tabular*}

\begin{minipage}{0.92\linewidth}
\footnotesize
\end{minipage}
\vspace{-2em}
\end{table}

\subsection{Ablation Study}

Table~\ref{tab:ablation_main} isolates the two components of \modelname on the 8B model.
Plain supervised crop augmentation has only a small effect on ScreenSpot-Pro, improving SFT from 59.8 to 60.5.
This confirms that the benefit does not come from adding cropped images alone.
Within RL, dynamic crop alone improves standard GRPO from 63.4 to 64.3, while adaptive supervision alone improves it to 64.8.
Combining both components gives the best result, 65.8 on ScreenSpot-Pro and 95.5 on ScreenSpot-V2, showing that view construction and self-verified anchoring address complementary failure modes.

\paragraph{Supervision strategy.}
Table~\ref{tab:ablation_supervision} compares different anchor-supervision strategies under the GRPO + multi-view rollout setting.
Simply adding a normalized oracle anchor without gating severely hurts performance,
dropping ScreenSpot-Pro from 64.3 to 57.8.
This is because many early groups are all-zero: all model rollouts receive reward 0,
so the model-only baseline has $\mu_{\mathrm{m}}=0$ and $\sigma_{\mathrm{m}}=0$.
If the ungated oracle anchor is still assigned reward 1, its normalized advantage becomes
approximately $1/\epsilon$, producing an excessively large supervised update on
unverified examples.
Constant SFT avoids this advantage explosion, but it applies the oracle signal
unconditionally and brings only marginal gain.
In contrast, our gated anchor activates only when the group is self-verified by at
least one maximum-reward model rollout.
Thus, all-zero groups receive no oracle update, while verified groups still benefit
from a positive stabilizing signal.
This yields the best results, 95.5 on ScreenSpot-V2 and 65.8 on ScreenSpot-Pro.

\subsection{Crop Perturbation Diagnostic}

Finally,
we conduct a training diagnostic to measure sensitivity to crop perturbations.
We sample 250 training examples and, for each, evaluate the original view with eight randomly cropped target-preserving views, using the same crop procedure as training.
We report original and crop-view accuracy, worst-view accuracy, view consistency rate (VCR), and prediction flip rate, where VCR measures prediction consistency for the same query across image inputs.
For the exact computation of VCR and prediction flip rate, see Appendix~\ref{app:view_consistent_group_rollout}.
Table~\ref{tab:view_setting_comparison} shows that \modelname improves both average accuracy and cross-view stability.
Compared with standard GRPO, \modelname raises crop accuracy from 93.00\% to 96.25\%, worst-view accuracy from 87.63\% to 92.42\%, and VCR from 88.38\% to 90.40\%.
It also reduces the prediction flip rate from 8.31\% to 5.80\%.

\section{Conclusion}
We presented \modelname, a view-consistent self-verified training framework that uses target-preserving cropped views, exact coordinate remapping, and a self-verified oracle anchor to better align RL with GUI grounding. Across benchmarks, scales, and backbones, \modelname improves accuracy and crop robustness, suggesting that view-consistent group construction is an effective training signal for GUI grounding.

\section*{Limitations}

\modelname is designed for actionable GUI grounding tasks whose supervision can be verified by coordinate-format rewards.
For datasets that mix actionable instructions with refusal-style examples, the anchor mechanism should be applied selectively rather than uniformly.
In our implementation, oracle coordinate anchors are used only for actionable samples with valid target boxes, while refusal or non-actionable samples can be routed to a separate training objective or excluded from anchor activation.
Equivalently, the anchor gate can be extended with task-type checks, so that refusal examples do not introduce non-coordinate anchor sequences into the cross-view group.
This suggests that mixed actionable/refusal datasets are compatible with the framework, but require refusal-aware routing or reward design to avoid conflating coordinate grounding with response-style learning.

\modelname also introduces additional optimization sensitivity through view-consistent cropping.
Although cross-view grouping improves reward diversity, overly aggressive cropping can increase variance in the RL update, especially when the crop probability or the number of cropped views is large.
We therefore use a conservative augmentation schedule: we reduce $p_{\mathrm{crop}}$, limit the number of crop views per group, and retain pass-through original views to stabilize training and reduce train--test mismatch.
These choices make the point-in-box reward reliable throughout training while preserving the robustness benefits of cross-view comparison.
Future work may further automate this schedule with adaptive crop probability, dynamic KL control, and format-aware early stopping.

\FloatBarrier
\bibliography{main}

\FloatBarrier
\appendix
\section{Appendix}

\subsection{Training Details}
The key optimization and training hyperparameters for the \modelname experiments are summarized in Table~\ref{tab:hyperparams}.
\begin{table}[h]
	\centering
	\begin{tabular}{@{}ll@{}}
		\hline
		\textbf{Hyperparameter}         & \textbf{Value}      \\
		\hline
		deepspeed                       & ZeRO-3              \\
		freeze\_vision\_modules         & true                \\
		max\_prompt\_length             & 4096                \\
		num\_generations                & 8                   \\
		per\_device\_train\_batch\_size & 8                   \\
		gradient\_accumulation\_steps   & 4                   \\
		bf16                            & true                \\
		torch\_dtype                    & bfloat16            \\
		data\_seed                      & 42                  \\
		gradient\_checkpointing         & true                \\
		attn\_implementation            & flash\_attention\_2 \\
		num\_train\_epochs              & 100                 \\
		learning\_rate                  & 1e-6                \\
		$\beta$                         & 0.04                \\
		logging\_steps                  & 1                   \\
		\hline
	\end{tabular}
	\caption{
		Hyperparameter settings used in the \modelname training experiments.
	}
	\vspace{-1em}
	\label{tab:hyperparams}
\end{table}
\subsection{Implementation Details of View-Consistent Group Rollout}
\label{app:view_consistent_group_rollout}
Our implementation follows the target-preserving crop construction described in the main method.
For each training example, we first decide whether to use view augmentation by a deterministic index-seeded Bernoulli sampler: 80\% of examples use dynamic crops, while 20\% remain pass-through examples with repeated original views.
This pass-through branch keeps standard full-screen layouts in the training distribution.

For augmented examples, let the original image size be $W\times H$, and let the normalized ground-truth box be
$B=(x_1,y_1,x_2,y_2)$.
We convert it to pixel coordinates as
\[
	B^{\mathrm{px}}=(x_1W,y_1H,x_2W,y_2H).
\]
For each cropped view $k$, the initial crop size is set to
$(w_k,h_k)=(0.9W,\,0.9H)$.
If the crop width or height is smaller than the target box itself, that dimension is enlarged to the box size and clipped by the image boundary.
This corner-case handling guarantees that the target UI element is never truncated, even for large elements or narrow screenshots.
Each crop window is then sampled from the feasible range of top-left locations that fully contain $B^{\mathrm{px}}$, ensuring target preservation across all augmented views.

We then compute the feasible integer range for the crop's top-left corner so that the full target box is inside the crop:
\begin{equation}
	\begin{aligned}
		l & \in [\max(0, x_2W-w), \min(W-w, x_1W)],  \\
		t & \in [\max(0, y_2H-h), \min(H-h, y_1H)] .
	\end{aligned}
\end{equation}
The implementation samples eight independent crop windows from these feasible ranges, matching \texttt{num\_generations}=8.
For each crop, the target box is remapped exactly into the crop coordinate frame,
\begin{equation}
	\begin{aligned}
		B'=\bigl( &
		\frac{x_1W-l}{w},
		\frac{y_1H-t}{h}, \\
		          &
		\frac{x_2W-l}{w},
		\frac{y_2H-t}{h}
		\bigr),
	\end{aligned}
\end{equation}
clipped to $[0,1]$, and finally converted to the model's 0--1000 coordinate system.
Each cropped image is resized again with \texttt{smart\_resize} before being passed to the VLM, but the supervision coordinates are computed from the pre-resize crop geometry, so the rollout group remains view-consistent under the crop-induced coordinate transform.

\paragraph{Metric definitions for Table~\ref{tab:view_setting_comparison}.}
For each instance, let $c_0$ denote the correctness on the original view and
$c_m$ denote the correctness on crop view $m$, $m=1,\ldots,8$.
We define the prediction flip rate as
\[
	\frac{1}{8}\sum_{m=1}^{8}\mathbb{I}[c_m \ne c_0],
\]
averaged over instances.

VCR is the fraction of instances for which all nine views have the same correctness label:
\[
	\mathbb{I}[c_0=c_1=\cdots=c_8].
\]

\subsection{Additional Ablation and Analysis}

\begin{table}[!ht]
	\centering
	\scriptsize
	\setlength{\tabcolsep}{4pt}
	\renewcommand{\arraystretch}{0.95}
	\caption{
		\textbf{Ablation study on the number of oracle anchors.}
		The shaded row indicates the default single-anchor setting.
	}
	\label{tab:ablation_num_supervision}
	\begin{tabularx}{\linewidth}{
			>{\raggedright\arraybackslash\hsize=1.2\hsize\linewidth=\hsize}X
			*{2}{>{\centering\arraybackslash\hsize=0.9\hsize\linewidth=\hsize}X}}
		\toprule
		\textbf{Anchors} & \textbf{SSV2} & \textbf{SSPro} \\
		\midrule
		\rowcolor{gray!10}
		1                & 95.5          & 65.8           \\
		2                & 95.3             & 65.0           \\
		4                & 95.2          & 64.0           \\
		\bottomrule
	\end{tabularx}
\end{table}%

\paragraph{Additional Ablation: Number of Oracle Anchors}
Table~\ref{tab:ablation_num_supervision} shows that one oracle anchor is sufficient and preferable.
Increasing the number of supervised completions from 1 to 2 or 4 reduces ScreenSpot-Pro accuracy from 65.8 to 65.0 and 64.0.
This matches the role of the anchor in Section~3: it should provide a positive direction for difficult groups while keeping most sequences in the training block model-generated.
Too many anchors shift the objective toward fixed supervised imitation and weaken the group-relative exploration signal.

\begin{table}[!ht]
	\centering
	\scriptsize
	\setlength{\tabcolsep}{4pt}
	\renewcommand{\arraystretch}{0.95}
	\caption{
		\textbf{Ablation study on the oracle-anchor point choice} (Qwen3-VL-8B).
	}
	\label{tab:ablation_anchor_point}
	\begin{tabularx}{\linewidth}{
			>{\raggedright\arraybackslash\hsize=1.35\hsize\linewidth=\hsize}X
			*{2}{>{\centering\arraybackslash\hsize=0.825\hsize\linewidth=\hsize}X}}
		\toprule
		\textbf{Oracle Anchor Point} & \textbf{SSV2} & \textbf{SSPro} \\
		\midrule
		\rowcolor{gray!10}
		Box center                   & 95.5          & 65.8           \\
		Random point in GT box       & 95.2          & 64.8           \\
		\bottomrule
	\end{tabularx}
	\vspace{-1.5em}
\end{table}%

\paragraph{Additional Ablation: Oracle-anchor point choice.}
Table~\ref{tab:ablation_anchor_point} studies whether the oracle anchor is sensitive to using the ground-truth box center.
We compare the default center-point anchor against an anchor sampled uniformly from inside the ground-truth bounding box, using the same Qwen3-VL-8B setting.
The random-point variant reaches 95.2 on ScreenSpot-V2 and 64.8 on ScreenSpot-Pro, close to but below the default center-point anchor at 95.5 and 65.8.
This suggests that the stabilizing effect does not rely exclusively on the exact box center being the most visually salient point, since any point inside the target box remains a valid click location.
At the same time, the deterministic center point avoids injecting additional target-coordinate noise and gives the best ScreenSpot-Pro accuracy, so we use it as the default oracle anchor.

\begin{table}[!ht]
	\centering
	\footnotesize
	\setlength{\tabcolsep}{4pt}
	\renewcommand{\arraystretch}{0.95}
	\caption{
		\textbf{Ablation study on the number of target-preserving views.}
		The shaded row indicates the default setting used by \modelname.
	}
	\label{tab:ablation_view}
	\begin{tabularx}{\linewidth}{
			>{\raggedright\arraybackslash\hsize=1.2\hsize\linewidth=\hsize}X
			*{2}{>{\centering\arraybackslash\hsize=0.9\hsize\linewidth=\hsize}X}}
		\toprule
		\textbf{Views} & \textbf{SSV2} & \textbf{SSPro} \\
		\midrule
		$K{=}1$        & 95.3          & 64.8           \\
		$K{=}2$        & 95.2          & 64.8           \\
		$K{=}4$        & 95.5          & 64.4           \\
		\rowcolor{gray!10}
		$K{=}8$        & 95.5          & 65.8           \\
		\bottomrule
	\end{tabularx}
\end{table}%

\paragraph{Additional Ablation: Number of views.}
Table~\ref{tab:ablation_view} studies the number of target-preserving views.
Using more views increases the diversity of geometric contexts compared with the single-view setting, but the effect is not strictly monotonic because some crops can remove useful context or introduce harder visual layouts.
We use $K{=}8$ in the main experiments because it obtains the best ScreenSpot-Pro accuracy, 65.8, while matching the GRPO group size used in our method.

\begin{table}[!ht]
	\centering
	\footnotesize
	\setlength{\tabcolsep}{3pt}
	\renewcommand{\arraystretch}{0.95}
	\vspace{-1em}
	\caption{
		\textbf{Ablation study on crop strategies.}
		Dynamic crop provides the strongest ScreenSpot-Pro accuracy.
	}
	\label{tab:ablation_crop_strategy}%
	\begin{tabularx}{\linewidth}{
			>{\raggedright\arraybackslash\hsize=1.35\hsize\linewidth=\hsize}X
			*{2}{>{\centering\arraybackslash\hsize=0.825\hsize\linewidth=\hsize}X}}
		\toprule
		\textbf{Crop Strategy}      & \textbf{SSV2} & \textbf{SSPro} \\
		\midrule
		No crop                     & 95.3          & 64.8          \\
		Cross-offset crop           & 95.7          & 65.0          \\
		\rowcolor{gray!10}
		Dynamic crop                & 95.5          & 65.8          \\
		\bottomrule
	\end{tabularx}
	\vspace{-1.5em}
\end{table}%

\paragraph{Additional Ablation: Crop strategy.}
Table~\ref{tab:ablation_crop_strategy} compares different crop constructions.
Cross-offset cropping gives the highest ScreenSpot-V2 number, but dynamic cropping performs best on ScreenSpot-Pro.
This is consistent with our goal: the dynamic target-preserving crop is most useful on challenging high-resolution interfaces where the model must remain correct under substantial view changes.

\begin{figure*}[t]
	\centering
	\begin{subfigure}[t]{0.32\textwidth}
		\centering
		\includegraphics[width=\linewidth]{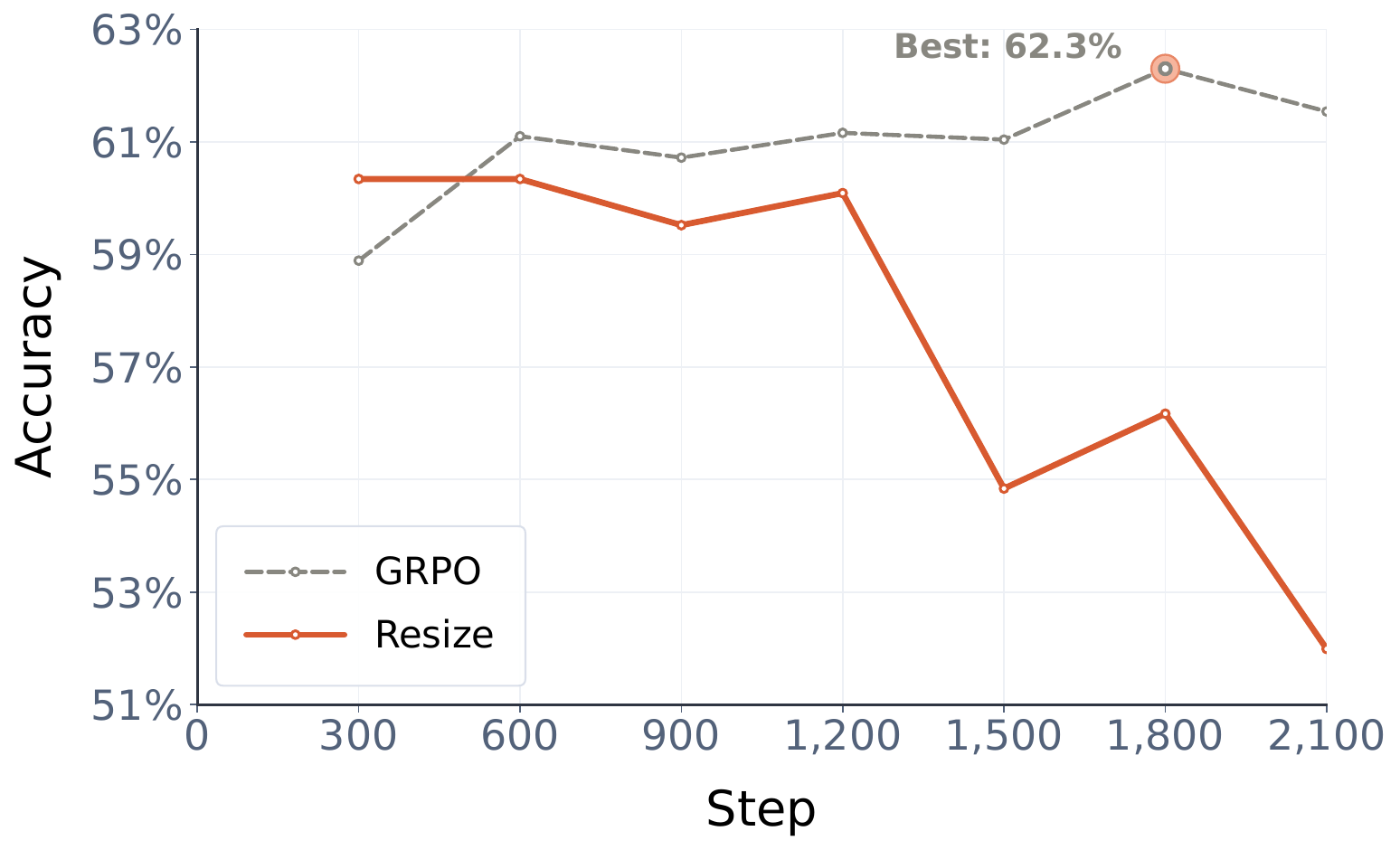}
		\caption{ScreenSpot-Pro accuracy $\uparrow$.}
		\label{fig:resize_sspro_accuracy}
	\end{subfigure}
	\hfill
	\begin{subfigure}[t]{0.32\textwidth}
		\centering
		\includegraphics[width=\linewidth]{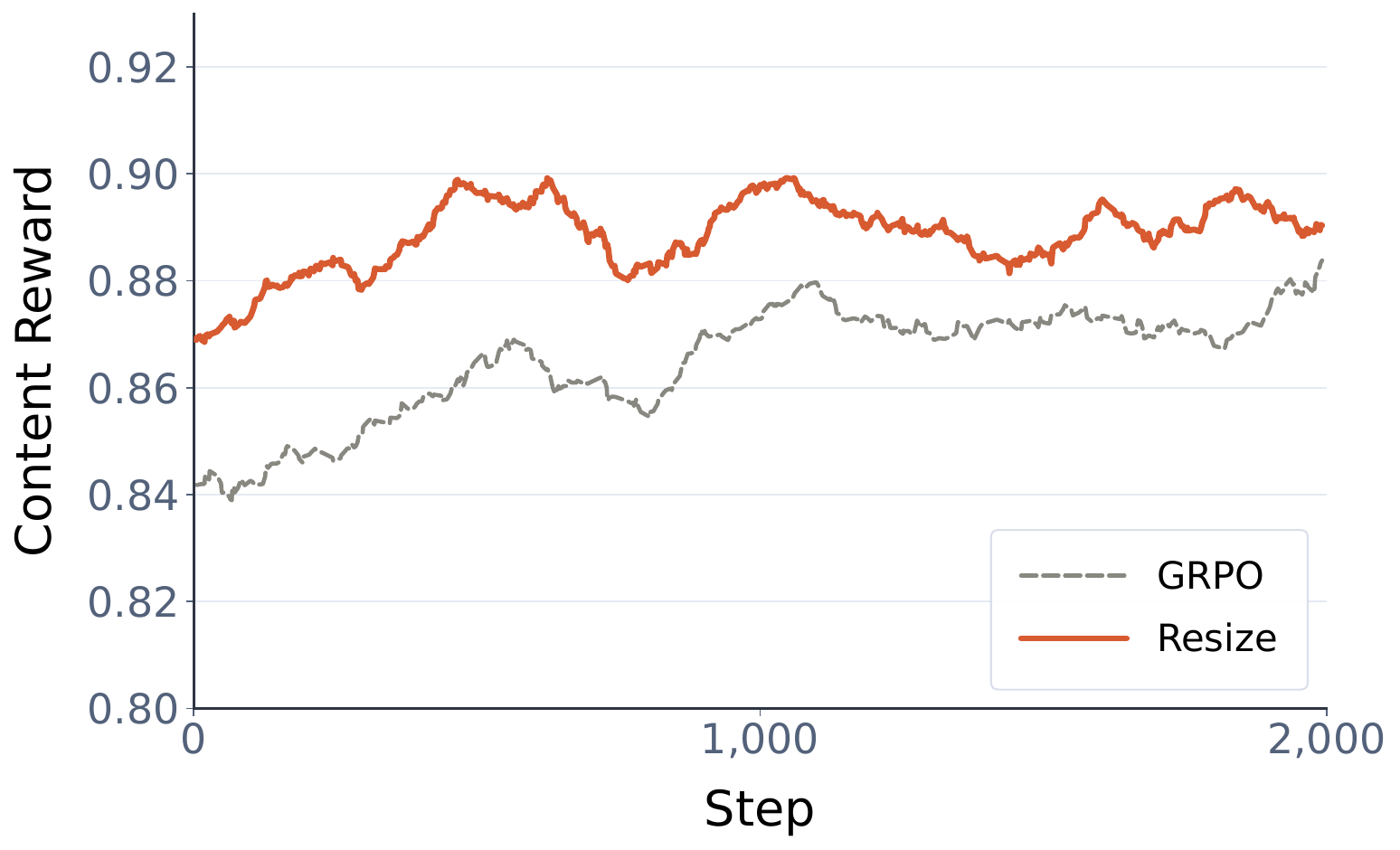}
		\caption{Content reward $\uparrow$.}
		\label{fig:resize_content_reward}
	\end{subfigure}
	\hfill
	\begin{subfigure}[t]{0.32\textwidth}
		\centering
		\includegraphics[width=\linewidth]{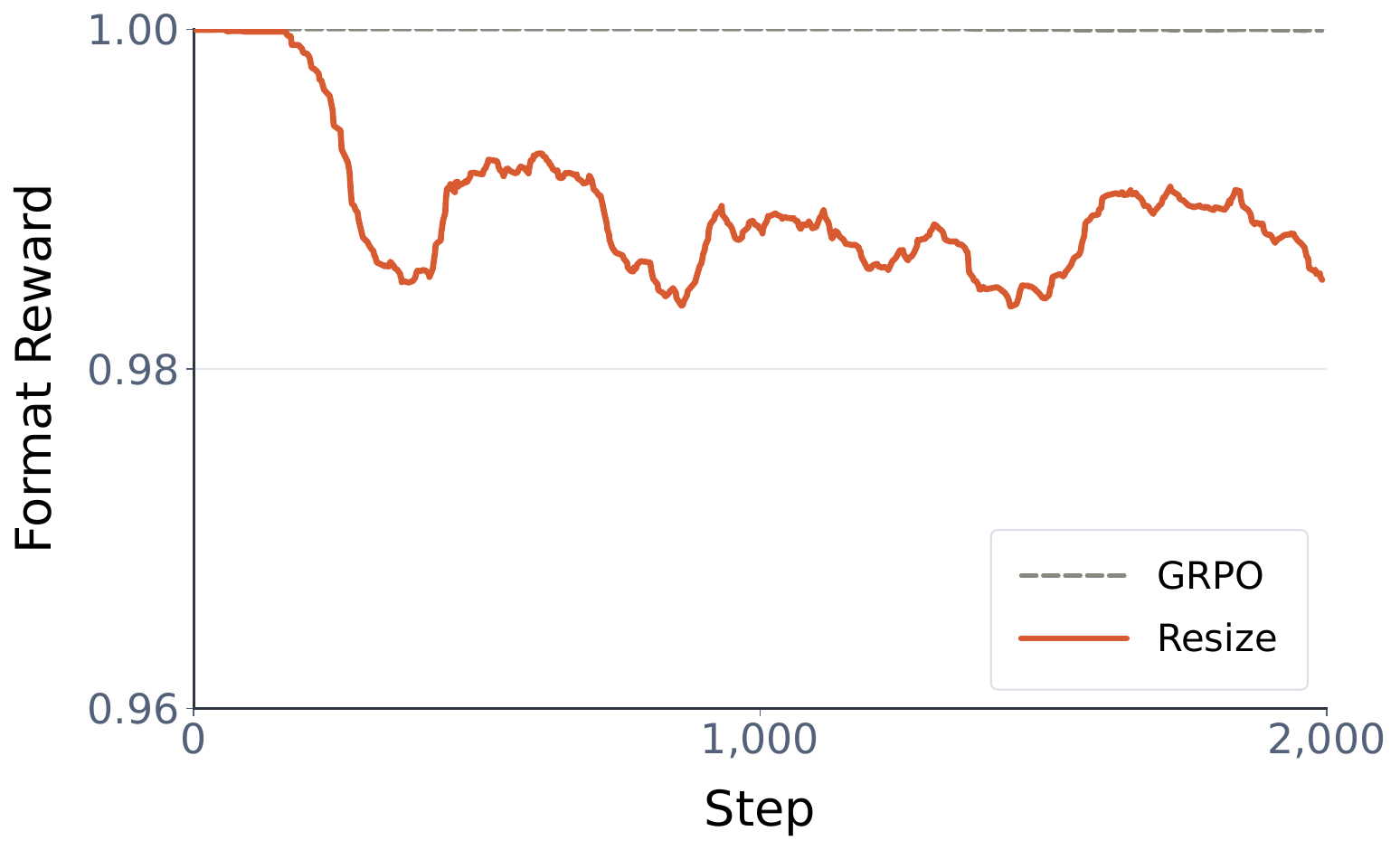}
		\caption{Format reward $\uparrow$.}
		\label{fig:resize_format_reward}
	\end{subfigure}
	\caption{\textbf{Resize strategy training dynamics and reward diagnostics.}
	}
	\label{fig:appendix_resize}
\end{figure*}

\paragraph{Additional Ablation: $p_{\mathrm{crop}}$.}

\begin{table}[!ht]
	\centering
	\footnotesize
	\setlength{\tabcolsep}{3pt}
	\renewcommand{\arraystretch}{0.95}
	\vspace{-1.5em}
	\caption{
		\textbf{Ablation study on $p_{\mathrm{crop}}$.}
	}
	\label{tab:ablation_p_crop}
	\begin{tabularx}{\linewidth}{
			>{\raggedright\arraybackslash\hsize=1.45\hsize\linewidth=\hsize}X
			>{\centering\arraybackslash\hsize=0.55\hsize\linewidth=\hsize}X
			>{\centering\arraybackslash\hsize=0.55\hsize\linewidth=\hsize}X}
		\toprule
		\textbf{$p_{\mathrm{crop}}$} & \textbf{SSV2} & \textbf{SSPro} \\
		\midrule
		1.0                          & 95.1          & 64.6           \\
		\rowcolor{gray!10}
		0.8                          & 95.5          & 65.8           \\
		0.6                          & 95.6          & 65.2           \\
		\bottomrule
	\end{tabularx}
	\vspace{-1.5em}
\end{table}

Table~\ref{tab:ablation_p_crop} studies the probability of applying target-preserving crop augmentation during training.
As described in Section~\ref{sec:method}, $p_{\mathrm{crop}}$ controls the mixture between view-consistent cropped groups and pass-through full-screen groups.
A larger value exposes the policy to more crop-induced coordinate transformations, while a smaller value retains more original screenshots and reduces train--test mismatch.
The results show that using crops for every example is not optimal: setting $p_{\mathrm{crop}}{=}1.0$ lowers ScreenSpot-Pro accuracy to 64.6, suggesting that removing full-screen pass-through examples makes the RL update overly dominated by cropped views.
Reducing the probability to $p_{\mathrm{crop}}{=}0.6$ slightly improves ScreenSpot-V2 to 95.6 but weakens ScreenSpot-Pro to 65.2, indicating insufficient view diversity for the harder high-resolution benchmark.
We therefore use $p_{\mathrm{crop}}{=}0.8$ in the main experiments, which achieves the best ScreenSpot-Pro accuracy, 65.8, while maintaining strong ScreenSpot-V2 accuracy, 95.5.

\paragraph{Additional Ablation: Image processing strategy.}
We further study whether the benefit of view-consistent rollout comes from
using multiple image inputs in general, or specifically from changing the
target coordinates through view transformation.
To this end, we conduct an additional training-dynamics experiment
initialized from Qwen3-VL-4B.
We compare the GRPO baseline with a simpler multi-image resize strategy,
denoted as GRPO + multi-image resize.
In this variant, each GRPO group is constructed from multiple resized
versions of the same full screenshot.
Unlike target-preserving cropping, resizing changes the visual scale but
largely preserves the target element's relative coordinate in the normalized
image space.
Figure~\ref{fig:appendix_resize} reports the ScreenSpot-Pro accuracy,
content reward, and format reward during training. ScreenSpot-Pro accuracy is
evaluated every 300 steps starting from step 300.

Although multi-image resize slightly outperforms the GRPO baseline at the
first evaluation point, its performance quickly becomes unstable and then
degrades substantially. The baseline GRPO curve remains relatively stable,
ending at 61.54, whereas GRPO + multi-image resize drops from 60.34 at step
300 to 51.99 at step 2100. The final gap reaches 9.55 points.

\begin{table*}[!t]
	\centering
	\scriptsize
	\setlength{\tabcolsep}{2pt}
	\renewcommand{\arraystretch}{0.95}
	\caption{Performance comparison on the \textbf{ScreenSpot-Pro.}
	$^*$ denotes our evaluated results.
	}
	\label{tab:main_results_screenspot_pro}
	\resizebox{\textwidth}{!}{%
		\begin{tabular}{l*{12}{c}c}
			\toprule
			\multirow{2}{*}{\textbf{Model}}     & \multicolumn{2}{c}{\textbf{CAD}} & \multicolumn{2}{c}{\textbf{Dev.}} & \multicolumn{2}{c}{\textbf{Creative}} & \multicolumn{2}{c}{\textbf{Scientific}} & \multicolumn{2}{c}{\textbf{Office}} & \multicolumn{2}{c}{\textbf{OS}} & \multirow{2}{*}{\textbf{Avg.}}                                                                                        \\
			\cmidrule(lr){2-3}
			\cmidrule(lr){4-5}
			\cmidrule(lr){6-7}
			\cmidrule(lr){8-9}
			\cmidrule(lr){10-11}
			\cmidrule(lr){12-13}
			                                    & \textbf{Text}                    & \textbf{Icon}                     & \textbf{Text}                         & \textbf{Icon}                           & \textbf{Text}                       & \textbf{Icon}                   & \textbf{Text}                  & \textbf{Icon} & \textbf{Text} & \textbf{Icon} & \textbf{Text} & \textbf{Icon} &      \\
			\midrule
			\multicolumn{14}{l}{\color{darkred}{\textit{$\approx$4B}}}                                                                                                                                                                                                                                                                                                                                   \\
			Qwen3-VL-4B$^*$~\citep{Qwen3-VL}    & 53.3                             & 18.8                              & 78.6                                  & 31.7                                    & 70.7                                & 22.4                            & 75.7                           & 30.0          & 83.1          & 39.6          & 76.6          & 33.7          & 55.5 \\
			GRPO-4B                             & 65.5                             & 25.0                              & 79.9                                  & 40.0                                    & 71.2                                & 28.7                            & 83.3                           & 40.0          & 90.4          & 47.2          & 74.8          & 39.3          & 61.5 \\
			\rowcolor{gray!10}
			{\modelname-4B}                     & 64.0                             & 28.1                              & 81.1                                  & 45.2                                    & 72.7                                & 33.6                            & 81.9                           & 40.0          & 89.8          & 54.7          & 76.6          & 40.5          & 63.4 \\
			\rowcolor{gray!10}
			\textit{+ MVP}                      & 79.2                             & 51.6                              & 87.2                                  & 53.4                                    & 74.2                                & 48.3                            & 85.4                           & 50.0          & 92.1          & 73.6          & 78.5          & 61.8          & 71.8 \\
			\midrule
			\multicolumn{14}{l}{\color{darkred}{\textit{$\approx$8B}}}                                                                                                                                                                                                                                                                                                                                   \\
			UI-TARS-7B~\citep{ui-tars}          & 20.8                             & 9.4                               & 58.4                                  & 12.4                                    & 50.0                                & 9.1                             & 63.9                           & 31.8          & 63.3          & 20.8          & 30.8          & 16.9          & 35.7 \\
			Phi-Ground~\citep{phi_ground}       & 26.9                             & 17.2                              & 70.8                                  & 16.7                                    & 56.6                                & 13.3                            & 58.0                           & 29.1          & 76.4          & 44.0          & 55.1          & 25.8          & 43.2 \\
			GUI-Actor-7B~\citep{gui_actor}      & 47.7                             & 9.4                               & 59.1                                  & 15.9                                    & 59.6                                & 16.1                            & 70.1                           & 25.5          & 69.5          & 41.5          & 55.1          & 19.1          & 44.6 \\
			SE-GUI-7B~\citep{SE_GUI}            & 51.3                             & 14.1                              & 68.2                                  & 19.3                                    & 57.6                                & 9.1                             & 75.0                           & 28.2          & 78.5          & 43.4          & 49.5          & 25.8          & 47.2 \\
			GUI-G$^2$-7B~\citep{GUI_G2}         & 55.8                             & 12.5                              & 68.8                                  & 17.2                                    & 57.1                                & 15.4                            & 77.1                           & 24.5          & 74.0          & 32.7          & 57.9          & 21.3          & 47.5 \\
			OpenCUA-7B~\citep{opencua}          & -                                & -                                 & -                                     & -                                       & -                                   & -                               & -                              & -             & -             & -             & -             & -             & 50.0 \\
			GTA1-7B~\citep{GTA1}                & 53.3                             & 17.2                              & 66.9                                  & 20.7                                    & 62.6                                & 18.9                            & 76.4                           & 31.8          & 82.5          & 50.9          & 48.6          & 25.9          & 50.1 \\
			UI-Venus-7B~\citep{ui_venus}        & 60.4                             & 21.9                              & 74.7                                  & 24.1                                    & 63.1                                & 14.7                            & 76.4                           & 31.8          & 75.7          & 41.5          & 49.5          & 22.5          & 50.8 \\
			InfiGUI-G1-7B~\citep{infiguig1}     & 57.4                             & 23.4                              & 74.7                                  & 24.1                                    & 64.6                                & 18.2                            & 80.6                           & 31.8          & 75.7          & 39.6          & 57.0          & 29.2          & 51.9 \\
			GUI-Owl-7B~\citep{mobileagentv3}    & 64.5                             & 21.9                              & 76.6                                  & 31.0                                    & 59.6                                & 27.3                            & 79.1                           & 37.3          & 77.4          & 39.6          & 59.8          & 33.7          & 54.9 \\
			MAI-UI-8B~\citep{mai-ui}            & 72.6                             & 35.9                              & 83.8                                  & 52.4                                    & 76.3                                & 33.6                            & 79.9                           & 37.3          & {88.7}        & 60.4          & 76.6          & {49.4}        & 65.8 \\
			Qwen3-VL-8B$^*$~\citep{Qwen3-VL}    & 56.9                             & 10.9                              & 75.3                                  & 22.8                                    & 68.2                                & 16.1                            & 78.5                           & 32.7          & 80.8          & 39.6          & 71.0          & 20.2          & 52.7 \\
			GRPO-8B                             & 73.6                             & 31.2                              & 82.5                                  & 39.3                                    & 77.8                                & 28.0                            & 81.9                           & 40.0          & 88.7          & 49.1          & 73.8          & 40.4          & 63.4 \\
			\rowcolor{gray!10}
			{\modelname-8B}                     & 74.6                             & 34.4                              & 85.3                                  & 45.9                                    & 73.7                                & 29.4                            & 81.9                           & 40.9          & 91.5          & 58.5          & 76.6          & 42.7          & 65.8 \\
			\rowcolor{gray!10}
			\textit{+ MVP}                      & 81.7                             & 45.3                              & 89.1                                  & 58.2                                    & 76.3                                & 42.7                            & 86.1                           & 56.9          & 92.1          & 71.7          & 80.4          & 53.9          & 72.0 \\
			\midrule
			\multicolumn{14}{l}{\color{darkred}{\textit{$\geq$30B}}}                                                                                                                                                                                                                                                                                                                                     \\
			Qwen3-VL-32B$^*$~\citep{Qwen3-VL}   & 60.4                             & 28.1                              & 69.5                                  & 22.1                                    & 75.8                                & 25.2                            & 84.7                           & 25.5          & 85.9          & 43.4          & 62.6          & 15.7          & 54.9 \\
			OpenCUA-32B~\citep{opencua}         & -                                & -                                 & -                                     & -                                       & -                                   & -                               & -                              & -             & -             & -             & -             & -             & 55.3 \\
			GUI-Owl-32B~\citep{mobileagentv3}   & 62.4                             & 28.1                              & {84.4}                                & 39.3                                    & 65.2                                & 18.2                            & 82.6                           & 39.1          & 81.4          & 39.6          & 70.1          & 36.0          & 58.0 \\
			GTA1-32B~\citep{GTA1}               & 43.7                             & 23.4                              & 82.5                                  & 28.3                                    & 69.2                                & 14.7                            & 79.9                           & 31.8          & 80.8          & 43.4          & 70.1          & 32.6          & 63.6 \\
			MAI-UI-32B~\citep{mai-ui}           & 70.1                             & {45.3}                            & {86.4}                                & 40.7                                    & {82.8}                              & 37.8                            & {91.7}                         & 46.4          & {90.4}        & 71.7          & {78.5}        & 34.8          & 67.9 \\
			UGround-v1-72B~\citep{uground}      & 16.8                             & 4.7                               & 55.8                                  & 4.8                                     & 54.0                                & 10.5                            & 70.8                           & 22.7          & 61.0          & 18.9          & 40.2          & 7.9           & 34.5 \\
			UI-Tars-72B~\citep{ui-tars}         & 18.8                             & 12.5                              & 63.0                                  & 17.2                                    & 57.0                                & 15.4                            & 64.6                           & 20.9          & 63.3          & 26.4          & 42.1          & 15.7          & 38.1 \\
			UI-Venus-72B~\citep{ui-venus}       & 66.5                             & 29.7                              & 84.4                                  & 33.1                                    & 73.2                                & 30.8                            & 84.7                           & 42.7          & 83.1          & 60.4          & 75.7          & 36.0          & 61.9 \\
			Qwen3-VL-30A3B$^*$~\citep{Qwen3-VL} & 51.8                             & 15.6                              & 76.0                                  & 24.8                                    & 69.2                                & 20.3                            & 76.4                           & 27.3          & 80.8          & 37.7          & 75.7          & 38.2          & 53.7 \\
			GRPO-30A3B                          & 69.5                             & 29.7                              & 84.4                                  & 51.7                                    & 74.7                                & 33.6                            & 81.9                           & 40.0          & 89.3          & 52.8          & 83.2          & 53.9          & 65.9 \\
			\rowcolor{gray!10}
			{\modelname-30A3B}                  & 71.1                             & 32.8                              & 85.9                                  & 52.1                                    & 76.8                                & 35.7                            & 83.3                           & 38.2          & 87.0          & 56.6          & 78.5          & 50.6          & 67.0 \\
			\rowcolor{gray!10}
			\textit{+ MVP}                      & 82.7                             & 45.3                              & 89.7                                  & 58.2                                    & 79.8                                & 51.8                            & 86.8                           & 49.1          & 92.1          & 71.7          & 83.2          & 60.7          & 74.1 \\
			\bottomrule
		\end{tabular}%
	}
	\vspace{-0.8em}
\end{table*}

We hypothesize that this failure is caused by the lack of coordinate
variation in the resized views. Resizing the full screenshot changes the
visual scale and image tokenization, but it does not change the target
element's relative location in the normalized coordinate system. Therefore,
different resized inputs in the same GRPO group correspond to nearly the
same correct coordinate output. In this case, the group mainly contains
multiple scale variants with an unchanged answer, so the model is not forced
to learn how the output coordinate should transform with the input view.

In contrast, target-preserving cropping changes the target element's
relative position in the cropped coordinate frame. For the same instruction,
different cropped views generally require different coordinate outputs after
exact box remapping. This creates a stronger cross-view grounding signal:
the model must localize the same semantic target under different visual
contexts and produce coordinates that are consistent with the crop-induced
coordinate transform. This explains why dynamic cropping provides useful
view-consistent supervision, whereas naive multi-image resize gives largely
redundant outputs and leads to unstable training.

\paragraph{Detailed ScreenSpot-Pro results.}
Table~\ref{tab:main_results_screenspot_pro} provides the category-level
ScreenSpot-Pro breakdown behind the aggregate SSPro numbers in
Table~\ref{tab:overview_results}. Across the 4B, 8B, and 30A3B settings,
\modelname improves the corresponding GRPO baseline by 1.3, 2.4, and 1.1
average points, respectively. Adding MVP further raises the averages to
71.8, 72.0, and 74.1, showing that the view-consistent training signal remains complementary to inference-time multi-view aggregation.

\end{document}